\documentclass[10pt,twocolumn,letterpaper]{article}

\usepackage[pagenumbers]{cvpr} %

\usepackage[dvipsnames]{xcolor}

\def\ParaHouse{ParaHome }

\newcommand{\vb}[1]{\mathbf{#1}}%
\newcommand{\mb}[1]{\mathbf{#1}} %

\newcommand{\bs}[1]{\boldsymbol{#1}}

\usepackage{times}
\usepackage{microtype}
\usepackage{epsfig}
\usepackage[skip=5pt]{caption}
\usepackage{float}
\usepackage{placeins}
\usepackage{color, colortbl}
\usepackage{stfloats}
\usepackage{enumitem}
\usepackage{tabularx}
\usepackage{xstring}
\usepackage{cuted} %
\usepackage{multirow}
\usepackage{xspace}
\usepackage{url}
\usepackage{subcaption}
\usepackage{mathtools}
\usepackage[hang,flushmargin]{footmisc}
\usepackage[normalem]{ulem}  %
\usepackage{graphicx}
\usepackage{amsmath}
\usepackage{amssymb}
\usepackage{comment}
\usepackage[title]{appendix}
\usepackage{cancel}
\usepackage{bbm}
\usepackage{subcaption}
\usepackage{afterpage}
\usepackage{algorithm}
\usepackage{algorithmic}
\usepackage{bm}
\usepackage{gensymb}
\usepackage{makecell}

\DeclareMathOperator*{\argmin}{arg\,min}

\newfloat{sepfigure}{p}{lop}
\floatname{sepfigure}{Figure}

\definecolor{R1color}{rgb}{0,0.56,0}
\newcommand{\R}[1]{{%
    \textbf{%
        \ifstrequal{#1}{1}{\textcolor[rgb]{0,0.56,0}{R#1}}{%
        \ifstrequal{#1}{2}{\textcolor{blue}{R#1}}{%
        \ifstrequal{#1}{3}{\textcolor{magenta}{R#1}}{%
        \ifstrequal{#1}{4}{\textcolor{teal}{R#1}}{%
                           \textcolor{cyan}{R#1}%
        }}}}%
    }%
}}

\usepackage{pifont}%
\newcommand{\cmark}{\textcolor{green!80!black}{\ding{51}}}
\newcommand{\xmark}{\textcolor{red}{\ding{55}}}

\usepackage{algorithm}
\usepackage{algorithmic}

\usepackage{bm}

\def\paperID{1227}
\def\confName{CVPR}
\def\confYear{2025}
\def\paperTitle{ParaHome: Parameterizing Everyday Home Activities\\ Towards 3D Generative Modeling of Human-Object Interactions}
\def\authorBlock{
    Jeonghwan Kim\thanks{Equal contribution} \qquad
    Jisoo Kim\footnotemark[1] \qquad
    {Jeonghyeon Na} \qquad
    {Hanbyul Joo}\\
    Seoul National University \\
    {\tt\small \{roastedpen,jlogkim,prom317,hbjoo\}@snu.ac.kr}\\
    {\tt\small \url{https://jlogkim.github.io/parahome}}     
}

\usepackage{indentfirst}

\definecolor{cvprblue}{rgb}{0.21,0.49,0.74}
\usepackage[pagebackref,breaklinks,colorlinks,citecolor=cvprblue]{hyperref}

\title{\paperTitle}

\author{\authorBlock}

\begin{document}
\maketitle

\begin{strip}
    \centering
    \includegraphics[width=\textwidth, trim={0cm 0.0cm 0cm 0cm},clip]{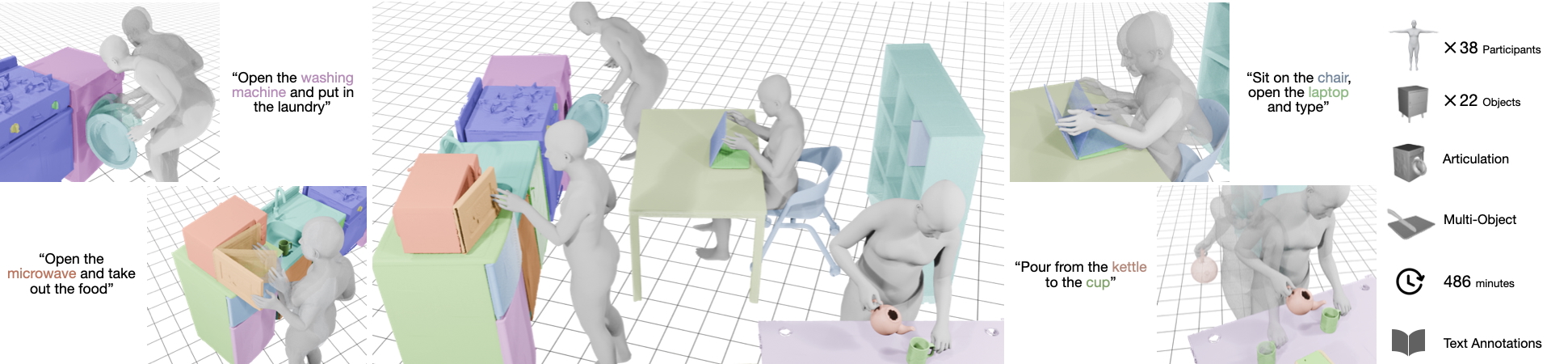}
    \captionof{figure}{Our system captures the detailed 3D movements of the human body, hands, and diverse objects, along with text descriptions.}
    \label{fig:teaser}
\end{strip}

\begin{abstract}
To enable machines to understand the way humans interact with the physical world in daily life, 3D interaction signals should be captured in natural settings, allowing people to engage with multiple objects in a range of sequential and casual manipulations.
To achieve this goal, we introduce our \ParaHouse system designed to capture dynamic 3D movements of humans and objects within a common home environment. 
Our system features a multi-view setup with 70 synchronized RGB cameras, along with wearable motion capture devices including an IMU-based body suit and hand motion capture gloves. 
By leveraging the \ParaHouse system, we collect a new human-object interaction dataset, including 486 minutes of sequences across 207 captures with 38 participants, offering advancements with three key aspects: (1) capturing body motion and dexterous hand manipulation motion alongside multiple objects within a contextual home environment; (2) encompassing sequential and concurrent manipulations paired with text descriptions; and (3) including articulated objects with multiple parts represented by 3D parameterized models. 
We present detailed design justifications for our system, and perform key generative modeling experiments to demonstrate the potential of our dataset. 

\end{abstract}

\section{Introduction}
\label{sec:intro}

Our daily routines involve interactions with various objects, where we perform complex, sequential, and dexterous motions. Machines, however, struggle to mimic these behaviors, as the connection between human action and environmental responses (e.g., pulling a refrigerator handle to open it) is challenging to model. The major obstacle is the lack of large-scale datasets capturing 3D human-object interactions (HOIs) in natural and casual settings, including the 3D motions of human bodies, hands, and objects within a common spatio-temporal space. 
Existing datasets cover limited aspects: capturing human motion without objects~\cite{gross2001cmu,sigal2010humaneva,ionescu2013human3,joo2015panoptic, mahmood2019amass}, capturing human body motions in a static environment without dexterous hand manipulations~\cite{wang2022humanise,shimada2022hulc,hassan2019resolving}, focusing on hand and object interactions in static postures without  motions~\cite{brahmbhatt2020contactpose,hampali2020honnotate,hasson2019learning,chao2021dexycb}, or considering relatively simple and atomic interactions such as grasping a single object~\cite{garcia2018first,taheri2020grab,zhang2023neuraldome,bhatnagar2022behave,li2023object, wang2023physhoi}.
While a few recent datasets try to capture both body motion and dexterous hand manipulation, they are often captured in a less natural setup (e.g., a simple table setup), containing limited action diversity~\cite{jiang2023full,fan2023arctic, zhan2024oakink2}.

To solve such fundamental challenges in collecting high-quality HOI data in a natural home environment, we present a novel capture system \textit{ParaHome}, along with a large-scale dataset including diverse and natural human-object interactions. Our system allows participants to freely interact with multiple objects in a room environment as shown in Fig.~\ref{fig:teaser}. Obtaining such signals is challenging, since the system should handle severe occlusions during interactions and multiple scales (big body movements across the room and subtle hand motion). As a solution, we design our \ParaHouse system by combining a multiview camera system equipped with 70 synchronized cameras, wearable IMU-based motion capture suit and gloves. The multi-camera system tracks the 3D movements of objects, capturing both rigid and articulated motions, as well as the global rigid transformations of the human body and hands in the camera space. The human body and hand motions are captured using wearable motion capture equipments that are occlusion-free enabling occlusion-robust capture of human object interaction. As combining heterogeneous systems is challenging, we present multiple hardware and algorithmic solutions to robustly reconstruct human motions and object movements in a common 3D spatiotemporal space. Through experiments, we verify the robustness of our system, providing justifications of our design choices. 

Leveraging our \ParaHouse system, we collect a large-scale human-object interaction dataset in the home activity scenarios. Our dataset contains data from 38 participants, 22 objects, 208 captures, and 486 minutes of sequences in clock time, which are publicly available. We design the capture scenario to reflect various natural activities that are commonly observed in our daily lives, where each participant performs sequential and casual interactions with multiple objects concurrently, in their own unique styles given a common verbal guidance in each capture. The resulting data captures detailed spatio-temporal relations between humans and multiple objects within 3D parametric spaces, where we provide corresponding 3D human mesh model (SMPL-X)~\cite{pavlakos2019expressive}, object-specific parametric models with their contact cues, and descriptive text annotaions, as shown in Fig.~\ref{fig:teaser}.

Our newly captured dataset provides new research opportunities for modeling the correlations between human motions and 3D object movements in natural HOI scenarios. To demonstrate this, we formulate our generative modeling pipeline in the context of motion synthesis, and explore two possible example problems: text-conditioned motion synthesis and object-guided motion synthesis.

\section{Related Work}
\label{sec:related}

\noindent \textbf{3D Motion Capture System and HOI Datasets.}
Pioneered by the work of Kanade et al.~\cite{Kanade-1997}, several systems propose to reconstruct human behaviors~\cite{Matusik-2000,Matsuyama-2002,Gross-2003,Petit-2009} with multiview capture system. With such system, early approaches pursue to capture human body motions markerlessly %
~\cite{Gall-09, Gavrila-96, Cheung-05,Plankers-03, Bregler-04, Kehl-06, Corazza-10, Vlasic-08, Brox-10, Stoll-11, deAguiar-2008, Vlasic-2008, Furukawa-2008}. %
For more robust human motion tracking, some works combined IMU sensors into their system ~\cite{guzov2021human,lee2024mocap,dai2022hsc4d,kaufmann2023emdb,liang2023hybridcap,pan2023fusing}. 

Such multi-view systems are used with additional cues to capture hand-object interactions to deal with issues such as severe occlusions or smaller scale of hands.
For example, some methods use RGB-D images with the optimization techniques

~\cite{hampali2020honnotate,hampali2022keypoint}, manual annotation~\cite{chao2021dexycb,zimmermann2019freihand}, or pretrained models~\cite{sener2022assembly101}. Some other methods use marker-based mocap systems~\cite{han2018online}, magnetic sensors~\cite{garcia2018first}, or mocap gloves~\cite{delpreto2022actionsense}, and some rely on synthetic data to model the interactions~\cite{hasson2019learning,jian2023affordpose}.
Other than hand-object interactions, body-object interactions are captured with manual annotations~\cite{bhatnagar2022behave}, IMUs combined with RGB data to track objects\cite{zhang2024hoi}, IMUs for body pose and RGB for tracking body with object~\cite{zhang2022couch,pons2023interaction}. 
Considering both body and hands, scale difference is a critical issue. 
As a resolution, methods using depth estimation and segmentation~\cite{wang2023physhoi}, mocap systems~\cite{taheri2020grab,li2023object}, multi-view cameras~\cite{zhang2023neuraldome,zhang2023ins}, and RGB-D setups~\cite{huang2022intercap} are used. To account for articulated objects rather than rigid objects,
egocentric RGB data is used with manual annotation to compute hand and object pose~\cite{liu2022hoi4d}. Some approaches use multi-view setup with markers attached to the surface of body and object to capture interaction with sittable object~\cite{jiang2023full} and others ~\cite{fan2023arctic,zhan2024oakink2,liu2024taco,zakour2024adl4d} use table setting to restrict body movements or set a placeholder in the synthetic environment~\cite{jiang2024scaling}. We summarize the key aspects of existing datasets and ours in supp. mat.

\noindent \textbf{Human Motion Synthesis.}  Various approaches are proposed to reconstruct static hand-object interactions using dataset-based regression~\cite{doosti2020hope}, spatio-temporal consistency~\cite{liu2021semi}, and utilize prior knowledge of hand-object contact~\cite{grady2021contactopt,cao2021reconstructing} or interaction fields~\cite{zhou2022toch,yang2021cpf}. Such methods have been extended to include body and rigid object poses via visibility cues~\cite{xie2023visibility,xie2024intertrack,wang2022reconstructing}.
On top of that, recent approaches use an additional condition to generate human body motion, either text annotations~\cite{petrovich2022temos,guo2022generating,tevet2022human,jome2023motionscript,zhang2022motiondiffuse} or scene contexts~\cite{cao2020long,wang2021synthesizing,wang2020motion,hassan2021samp,wang2022towards,liu2023revisit,mir2023generating,wang2022humanise,yi2024tesmo}. Some methods rather focus on object manipulation~\cite{li2023object,kulkarni2023nifty,xu2023interdiff,hassan2023synthesizing,diller2023cg,wang2023physhoi,li2023controllable} and sometimes conditioned on text descriptions~\cite{peng2023hoi,yang2024f,xu2024interdreamer} without getting hand pose.
Several methods focus on getting plausible hands by exploring to
integrate hand pose into rigid object manipulation by generating grasp poses or motions given a target object trajectory~\cite{corona2020ganhand,turpin2022grasp,christen2022d,zhou2024gears}, dynamic manipulation with hands and object trajectories~\cite{zhang2021manipnet}, and text-based hand manipulation with rigid objects~\cite{christen2024diffh2o}.
Considering both body and hand, some approaches %
address body motion generation for grasping rigid objects~\cite{taheri2022goal,zheng2023coop,tendulkar2023flex,luo2024grasping}. Moving beyond simple grasping, some methods generate full-body motion and object interactions conditioned with action label~\cite{ghosh2023imos}, while others use correlations to generate object positions or scenes~\cite{petrov2023object,nie2022pose2room,yi2023mime}.
More recently, methods have synthesized hand motions conditioned on object states, including articulated motions~\cite{zhang2023artigrasp,zhang2024manidext,zheng2023cams}, or generate hand-object interactions based on text input~\cite{cha2024text2hoi}. 

\section{3D Parametric Home Capture System}
\label{sec:system}

\subsection{HOI Data Parameterization} To effectively model human-object interactions, we consider a parameterized 3D space that captures the nuanced relationship between human motions and object movements.
For the human parts, we focus on the motion of the whole body including dexterous hand movements as the essential components. Rigid objects are represented through 6 DoF rigid motions, and %
we also incorporate object-specific dynamics such as opening the laptop or turning the knobs of a gas stove. Formally, we represent the status of a human and the environment at time $t$ as $\mathbf{S}(t) = \{ \mathbf{S}_p(t), \mathbf{S}_e(t), \mathcal{T} \}$,
where $\mathbf{S}_p(t)$ is the current status of a human subject at time $t$, the environment status $\mathbf{S}_e(t)$ represents the current status of surrounding objects, and $\mathcal{T}$ denote a corresponding language description.
The human status $\mathbf{S}_p(t) = \{ \mathbf{S}_b(t), \mathbf{S}_{lh}(t), \mathbf{S}_{rh}(t)\}$ is composed of the body $\mathbf{S}_b(t)$, left-hand $\mathbf{S}_{lh}(t)$, and right-hand parameters $\mathbf{S}_{rh}(t)$, where each of them can be represented as mocap outputs via a global location and local joint orientations. 
The environment status is represented as $\mathbf{S}_e(t) = \{ \mathbf{S}^j_e(t)\}_{j=1}^N$, where $\mathbf{S}^j_e(t)$ is the status of the $j$-th object, assuming we consider $N$ different object instances. The object status $\mathbf{S}^j_e(t) = \{ \vb{l}^j_e(t), \boldsymbol{\theta}^j_e(t), \bs{\phi}^j_e(t)\}$ is represented by 3D translation $\vb{l}^j_e(t) \in \mathbb{R}^3$, 3D orientation  $\boldsymbol{\theta}^j_e(t) \in SO(3)$, and object specific parameters $\boldsymbol{\phi}^j_e(t)$. The length of $\bs{\phi}^j_e(t)$ varies based on the type and the dimension of movable parts of the objects (e.g. $\bs{\phi}^{\text{laptop}}_e(t)$ and  $\bs{\phi}^{\text{drawer}}_e(t)$ contain one and two part parameters each).

\subsection{Hardware System and Architecture}

\begin{figure*}[t!]
    \centering
    \includegraphics[width=\linewidth, trim={0cm 0.0cm 0cm 0cm},clip]{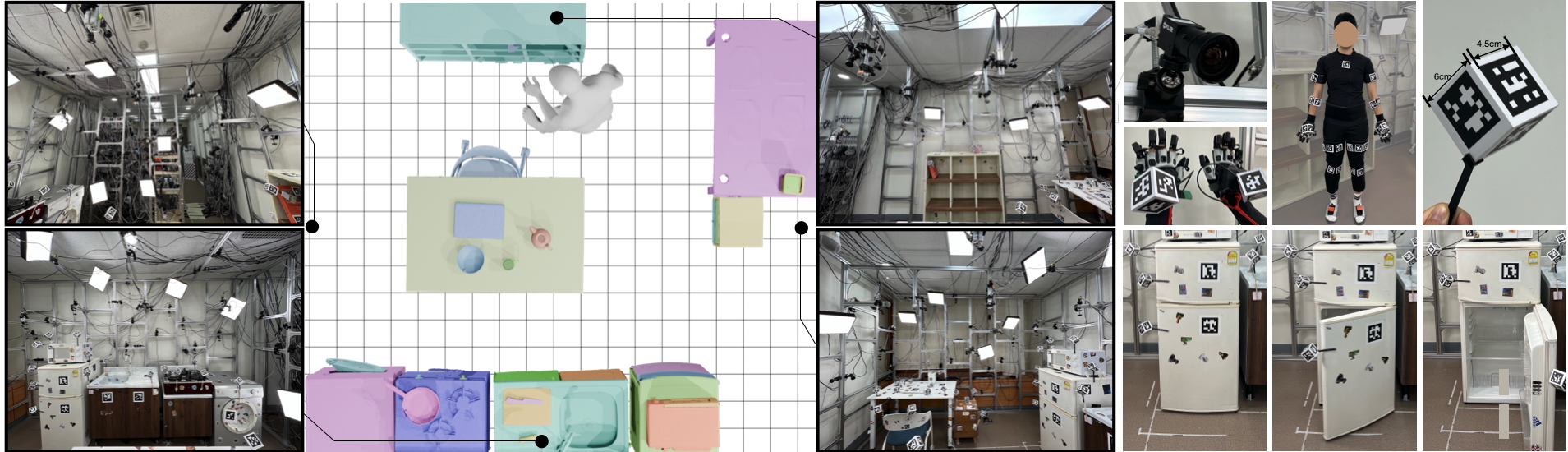}
    \caption{(Center) Reconstructed scene of \ParaHouse from top view. Pictures adjacent to the rendering were taken from the center of the room, headed towards the corresponding black dots in the scene. (Right) Pictures of RGB camera, IMU based motion capture devices with attached body markers and the 3D marker solution on an articulated object.}
    \label{fig:system_hardware}
\end{figure*}

Our \ParaHouse system encompasses an area of 12.4$m^2$, as shown in Fig.~\ref{fig:system_hardware}. 
To capture full body motion, subtle hand motions, and 3D object movements across the room setting, our system integrates a multi-camera system and an IMU-based wearable motion capture suit and gloves, leveraging their complementary strengths. 
To cover the entire volume of the room and reduce occlusion issues, we install 70 RGB industrial cameras. We use Xsens motion suit~\cite{xsens} and Manus hand gloves~\cite{manus} for wearable motion capture solutions. For further system details, see supp. mat..

\subsection{Modeling and Tracking 3D Objects}

\noindent \textbf{3D ArUco Marker Design and Placement.} %
To reliably track objects and their articulated
motions, we attach ArUco markers \cite{garrido2014automatic} to all faces of the 3D cube spanning 6$cm$ inspired by \cite{zoss2018empirical} as shown in Fig.~\ref{fig:system_hardware}. The major motivation of our 3D marker solution lies in its robust tracking advantage during complicated manipulation scenarios, where often objects are severely occluded by hand grasping. In particular, we find attaching markers on object surface as in previous approaches~\cite{fan2023arctic} is sub-optimal, often requiring laborious manual post-processing. We quantitatively demonstrate the strength and robustness of our 3D marker solution in Sec.~\ref{sec:sys_eval}.

\begin{figure}[tb]
    \centering
    \includegraphics[width=\columnwidth, trim={0cm 0.0cm 0cm 0cm},clip]{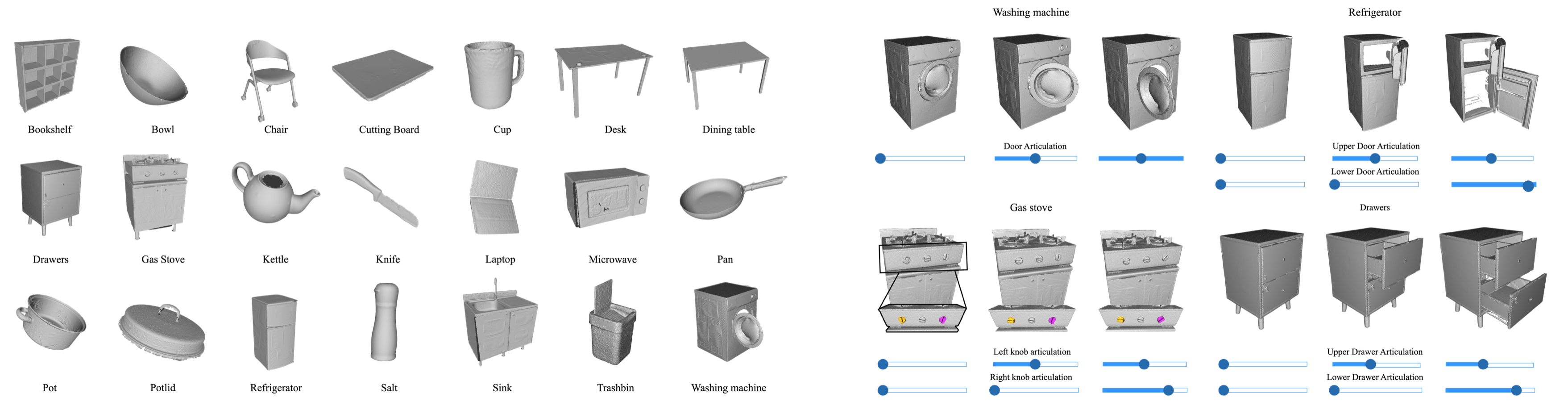}
    \caption{(Left) Scanned 3D models in \ParaHouse system. (Right) Articulation state of 3D models. Blue bars show the object-specific parameters $s_{e,i}^j(t)$ for each object part $i$. As $s_{e,i}^j(t)$ changes, corresponding parts of the objects show different articulation states}
    \label{fig:obj_combined}
    \vspace{-10pt}
\end{figure}

\noindent \textbf{Computing Transforms between Object and 3D Markers.}
We obtain high-quality 3D mesh of all objects placed in our system via off-the-shelf scanner and manual alignment. Our scanned objects are shown in Fig~\ref{fig:obj_combined}.
For better visibility, we attached one or multiple 3D ArUco markers and flat-style markers to each object part. Then, objects are tracked by 
detecting the ArUco markers in every frame. The position and orientation of each object are obtained as follows: %
\begin{equation}
    \mb{T}_{obj}(t) = \mb{T}_{mar \rightarrow obj} \mb{T}_{mar}(t)
    \label{eq:marker_to_obj}
\end{equation}
 where $\mb{T}_{mar \rightarrow obj}$ is a pre-computed fixed transformation from the marker to the object, and $\mb{T}_{mar}(t)$ is a transformation for the markers from the object canonical space to the current pose in the camera system space at capture time $t$. 
$\mb{T}_{mar \rightarrow obj}$ is computed by selecting several corresponding points between the object scan and the one in camera space.

\noindent \textbf{Modeling Object Articulations.}
Since the whole parts of an object are scanned as a single chunk, we manually separate each rigidly moving part and build a parameterized structure for articulated models.
We assume either a revolute or sliding joint. Both types of joints require articulation axis $\vb{a}_e \in \mathbb{R}^3$ and in case it is revolute, pivot point $\vb{p}_e \in \mathbb{R}^3$ additionally. Thus, we define the joint status of $j$-th object as $\bs{\phi}_e^j(t) = \{\bs{\tau}_i^j\}_{i=1}^n $ where $ \bs{\tau}_i^j = \{ \vb{a}_{e,i}^j, \vb{p}_{e,i}^j, s_{e,i}^j(t)\}$. Here, $n$ represents the number of parts, and $s_{e,i}^j(t) \in \mathbb{R}$ denotes a relative part state (either radian for revolute joint or meter for sliding joint) from part status in the object canonical space. Examples are shown in Fig.~\ref{fig:obj_combined}. See supp. mat. regarding the process of computing articulation axis $\vb{a}_e^j$ and pivot point $\vb{p}_e^j$.%

\noindent \textbf{Capturing 3D Object Motions.}
We track the status of each object at time $t$, $\mathbf{S}_e(t) = \{ \vb{l}_e(t), \bs{\theta}_e(t), \bs{\phi}_e(t)\}$, via our multi-view system.
The 3D rigid transformation (translation $\vb{l}_e(t)$ and 3D orientation  $\bs{\theta}_e(t)$) is computed by specifying the attached marker corners at each time $t$, triangulating N number of 2D marker points $\{\vb{m}_i(t)\}_{i=0}^N$ into 3D marker points $\{\vb{M}_i(t)\}_{i=0}^N$. Given the 3D marker corners $\{\vb{M}_i(t)\}_{i=0}^N$ and the corresponding corners $\{\vb{\hat{M}}_i\}_{i=0}^N$ in the object canonical space, we compute $\mb{T}_{mar}(t)$, via the Kabsch algorithm. Then, the object transformation $\mb{T}_{obj}(t)$ can be computed as Eq.~\ref{eq:marker_to_obj}. The object-specific dynamic status $\bs{\phi}_e(t)$ is similarly acquired via transformations from the 3D markers attached to the base part to those on the movable parts.
\subsection{Capturing 3D Human Motions}
\label{sec:Capturing_3D_Human_Motions}
It is challenging to leverage two heterogeneous systems, a multi-camera system and wearable motion suits and gloves, since they do not share a common spatial world coordinate. Specifically, wearable motion capture systems suffer from the drift issue in localizing the global root position, and more critically they employ an imperfect assumption of the body and hand skeleton scale that differs from the actual measurement. As such, directly transferring the output of wearable captures into the camera system space cannot fulfill our goal of capturing precise hand-object interactions. In this section, we present a method to spatially align two systems.

\noindent \textbf{Aligning Wearable Mocap in Multi-Camera System.} To spatially align the body motion capture with our multi-view camera system, we need to provide correspondences between the two systems. 
For this purpose, we attach 3 or 4 ArUco markers to each of 11 near-rigid body parts during body alignment capture(torso, hands, upper arms, lower arms, upper legs, lower legs), as shown in  Fig.~\ref{fig:system_hardware}.
We denote the four corner points of the $j$-th marker attached to the $i$-th body part as $\mathcal{M}^b_{i, j}\in \mathbb{R}^{4 \times 3}$, where the positions are defined in the local joint coordinate w.r.t the corresponding body part. %
Specifically, the goal of our alignment process is: (1) to obtain authentic body skeleton configuration $\mathcal{B}=\{\mathcal{O} \}$, which represents the offsets of the child joints from the parent joints, and (2) body marker locations $\mathcal{M}^b$. 
Given these parameters and the joint angle measurements provided by the mocap suit  $\theta^t \in \mathbb{R}^{23\times3}$ at time $t$, we can transform the body-attached markers into the person-centric coordinate denoted as $\mb{M}^b_{mocap}(t)$ via forward kinematics function $\mathcal{K}_b$ as: $\mb{M}^b_{mocap}(t) = \mathcal{K}_b(\mathcal{M}^b, \theta^t, \mathcal{B})$.
 At the same time, the body-attached markers can be reconstructed via our multi-view system, denoted as $\mb{M}_{cam}^b(t)$ defined in the camera system space. Then, the rigid transformation to transfer the mocap data into the camera system space can be computed with the marker correspondence: $T_{b}^{cam}(t) = T(\mb{M}^b_{mocap}(t),  \mb{M}_{cam}^b(t))$, where only visible markers in $\mb{M}_{cam}^b(t)$ are considered for the computation. 
 
\begin{figure}[tb]
    \centering
    \includegraphics[width=\linewidth, trim={0cm 0.0cm 0cm 0cm},clip]{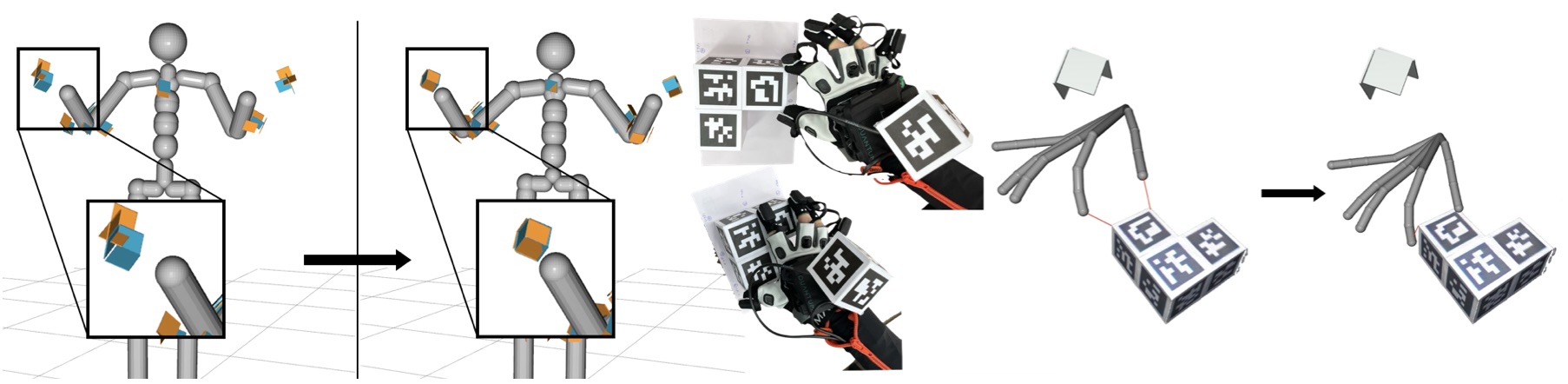}
    \caption{(Left) Before/After Body Calibration, Orange: forward kinematic output, Blue: RGB Triangulated Result (Right) Hand Calibration Protocol and Before/After Calibration Protocol
    }
    \label{fig:calibration}
    \vspace{-10pt}
\end{figure}

 Note that we can compute the rigid transformation as long as at least one body-attached marker is visible in $\mb{M}_{cam}^b(t)$, providing robustness to the marker occlusions in computing the global location of the actor. We perform such transformation for the entire frames of body alignment capture which consists of motions that encompass various body poses. Our objective function for $\mathcal{B}$ and $\mathcal{M}^b$ is defined as:
\begin{align} 
  \min_{\mathcal{M}^b, \mathcal{B}} \, \sum_{t=1}^{T}
{\lambda_{b}\mathcal{L}^t_{body} + \lambda_{f}\mathcal{L}^t_{foot}}.
  \label{eq:bodyoptim_loss_equation_sum}
\end{align}
$\mathcal{L}_{body}$ is the mean-squared error between corresponding markers. See supp. mat. for further details.
We use a constraint term $\mathcal{L}_{foot}$ to enforce the foot parts to be close to the ground without penetration or floating. The result of our body alignment process is shown in Fig.~\ref{fig:calibration}, showing the aligned markers after the optimization.

\noindent \textbf{Calibrating Hand Mocap.} 
The hand mocap outputs from the gloves also suffer from similar issues which are unknown skeleton lengths of the actual hand and localizing into the camera system space. Since a subtle error may cause a large deviation in the hand interaction scene, such issues are critical in obtaining high-quality data.
As a solution, we present a new hand alignment protocol using a calibration structure. We specify the ordered 3D corner vertices of the structure, $ C = \{\vb{c}_i \in \mathbb{R}^3 \}_{i=1}^{6}$, which we localize in camera system space via triangulation. 
During the alignment protocol, the participant touches the calibration structure's known corner locations with their fingertips, from which we can approximate the desired locations of fingertips.
Examples of hand calibration protocols and the hand calibration structure are shown in Fig.~\ref{fig:calibration} and our supplementary video as well. 

Similar to the body calibration, the goal of hand calibration is to obtain the authentic hand skeleton configuration $\mathcal{H}=\{\mathcal{S}^h\in \mathbb{R}^{20\times1},  \mathcal{O}^h\in \mathbb{R}^{20\times3}\}$ and positions of 3D markers in the local hand-centric coordinate $\mathcal{M}^h\in \mathbb{R}^{3\times4\times3}$, where $\mathcal{S}^h$ and $\mathcal{O}^h$ each denotes scales of hand skeleton and per-joint skeletal offsets. We perform an optimization: 
\begin{align} 
  \label{eq:handoptim_equation}
  \min_{\mathcal{M}^h, \mathcal{H}} \lambda_{t} L_{tip} +  \lambda_{w}\mathcal{L}_{wrist} +\lambda_{p}\mathcal{L}_{pen}.
\end{align}
$\mathcal{L}_{tip} = \sum_{k \in \mathbf{o}} \left \| T^k_h \mathcal{K}_h^{k}(\mathcal{H})_i-\vb{c}^k_j \right \|$ penalizes the difference between a fingertip and the corresponding corner of the calibration cube $c_j^k$ ($\mathbf{o}$ denotes the order of correspondences),  where $T^k_h$ is the hand-to-camera transformation at $k$-th time 
and $\mathcal{K}_h$ is the forward kinematics function to transform the $i$-th fingertip.
We add $\mathcal{L}_{wrist}$ to penalize the distance between the wrist positions from the body mocap and from the hand marker to enforce hands are rigidly connected to the body mocap. We also add $\mathcal{L}_{pen}$ to prevent fingertips from penetrating %
the calibration cube. See supp. mat. for details. 

\subsection{Post-Processing to Enhance Qualities}
As an advantage of the heterogeneous nature of our systems, we can leverage cues from both systems to reduce jitters and handle tracking failures. %

\noindent \textbf{Enhancing Hand Tracking.}
The global hand location estimated by the attached 3D marker may suffer from motion jitter or tracking failures due to occlusions or low camera visibility.  
Given two measurements, regarding the wrists from hand-attached markers and the body mocap, we perform a weighted interpolation to reduce motion jitters. In the case of tracking failure of the hand-attached 3D marker, we rely on the body motion from mocap since it's occlusion-free. We demonstrate the robustness of method in our experiments. See our supp. mat. for details.

\noindent \textbf{Enhancing Object Tracking.}
Although we use multiple cameras with our 3D marker system,  object tracking failure may still happen due to severe occlusions. As a solution, we leverage the body motion measurement from our system to enhance object tracking qualities.  
Specifically, we first infer whether an object is currently close to the human actor by checking the distances between human joint positions and the object's surface. If an object has been moved by the actor, we assume the object is near-rigidly connected to the close body joints, and apply the temporal transformation of the body joints to interpolate the missing object trajectories from the neighboring times. In practice, we apply this method for both reducing jitters and recovering the tracking failures. 

\begin{figure}[t!]
    \centering
    \includegraphics[width=\linewidth, trim={0cm 0.0cm 0cm 0cm},clip]{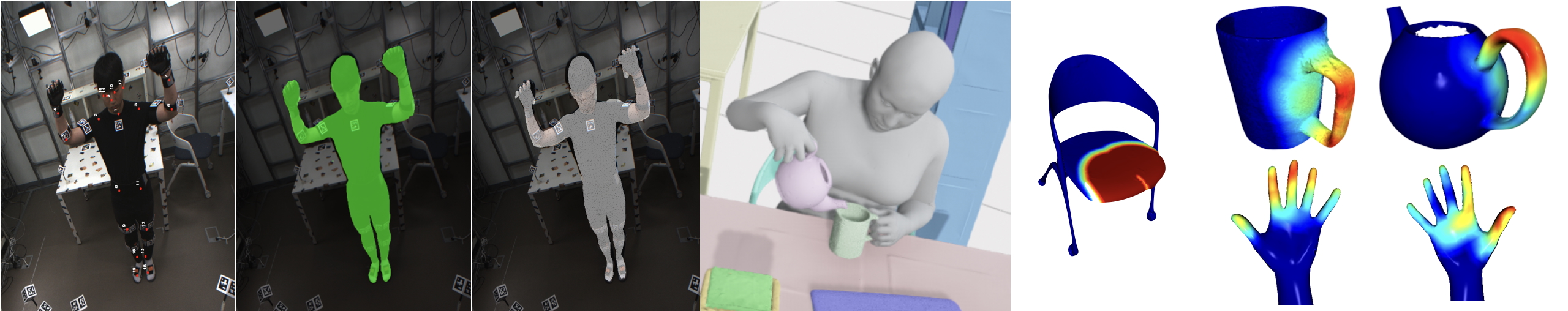}
    \caption{An example of SMPL-X shape parameter fitting. (Left) Projected keypoints, %
    mask and rendered SMPL-X with the optimized shape parameter. (Right) Rendered SDF 
    within 
    $5cm$ to visualize an affordance information using optimized SMPL-X.
    }
    \label{fig:smplx_fitting}
    \vspace{-10pt}
\end{figure}

\noindent \textbf{Fitting Human Body Model.} It is crucial to have 3D human mesh model aligned to each capture participant to reason about accurate contact between humans and the objects. %
We start by fitting SMPL-X~\cite{pavlakos2019expressive} shape parameter to each participant. Using HQ-SAM~\cite{sam_hq}, we initialize body segmentation masks and refine inaccurate pixels manually from 3 to 4 camera views. Then we optimize the parameters by aligning the rendered SMPL-X with the masks as well as projected joints. A result sample is shown in Fig.~\ref{fig:smplx_fitting}. For the pose parameter, directly computing SMPL-X pose parameter to fit into Xsens skeleton causes unnatural or inaccurate body motions even with optimized shape parameters due to different skeletal configurations between Xsens~\cite{xsens} and SMPL-X.
To enhance the fitting quality, we design a simple autoencoder %
to learn natural body motion manifold space using both AMASS~\cite{mahmood2019amass} and our Xsens body pose data. 
Using our trained model, we get an initial latent vector from Xsens body motion input, followed by further latent code optimization so that the decoded SMPL-X body pose fits to Xsens wrist, ankle, foots and figer tips. An example of fitting result is shown in Fig.~\ref{fig:smplx_fitting}. See the supp. mat. for details.

\begin{figure}[t!]
    \centering
    \includegraphics[width=\linewidth, trim={0cm 0.0cm 0cm 0cm},clip]{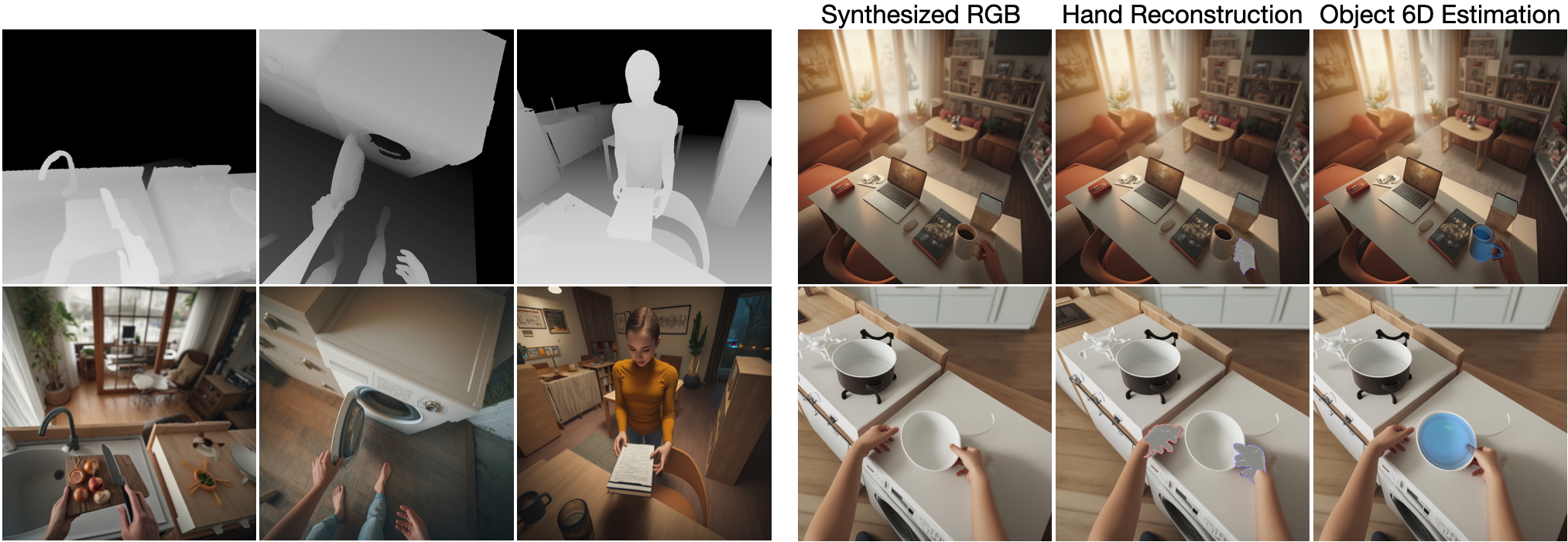}
    \caption{(Left) Examples of synthesized RGB images using ParaHome data. (upper) Rendered depth images of \ParaHouse data. (lower) Synthesized RGB image using text annotation, depth, 2D keypoints. (Right) HOI Reconstruction using synthesized RGB
    }
    \label{fig:synthetic_parahome}
    \vspace{-10pt}
\end{figure}

\noindent \textbf{Synthesizing realistic RGB.} 
Inspired by recent approaches Obman~\cite{hasson19_obman} and InterTrack~\cite{xie2024intertrack} which demonstrate the usefulness of synthetically produced RGB data paired with 3D GTs, we also provide corresponding synthetic RGB images for our \ParaHouse dataset by leveraging diffusion-based image synthesis model~\cite{flux}.
We render our 3D scenes from various viewpoints, including egocentric views, to produce depth maps and 2D body joint cues, which are used as the control inputs for synthesizing realistic RGB images with a ControlNet model~\cite{flux}.
We also find that incorporating our corresponding text annotations as an additional input for the ControlNet can further enhance the quality of the synthesis outputs. Examples are shown in Fig.~\ref{fig:synthetic_parahome}. 

To assess the realism and efficacy of the generated images, we perform a quantitative evaluation by applying off-the-shelf 3D estimation methods on our synthesized RGB as inputs to two tasks: RGBD-based 6D object pose estimation~\cite{zhang2025omni6d} and RGB-based 3D hand pose estimation~\cite{pavlakos2024reconstructing}. We compare these results to those from existing benchmarks (ROPE~\cite{zhang2025omni6d}, FreiHAND~\cite{zimmermann2019freihand}, and HO3D~\cite{hampali2020honnotate}), and demonstrate that recent state-of-the-art methods achieve similar patterns in performance. 
Notably, performance on our dataset is lower than on previous benchmarks, presumably due to our dataset’s more complex human-object interaction scenarios with mutual occlusion. Qualitative results are shown in Fig.~\ref{fig:synthetic_parahome} and quantitative results
are in supp. mat..

\section{Validating Capture System and Data} \label{sec:sys_eval}%

\subsection{System and Capture Evaluation} 

\begin{figure}[t!]%
    \centering
    \includegraphics[width=\linewidth, trim={0cm 0.0cm 0cm 0cm},clip]{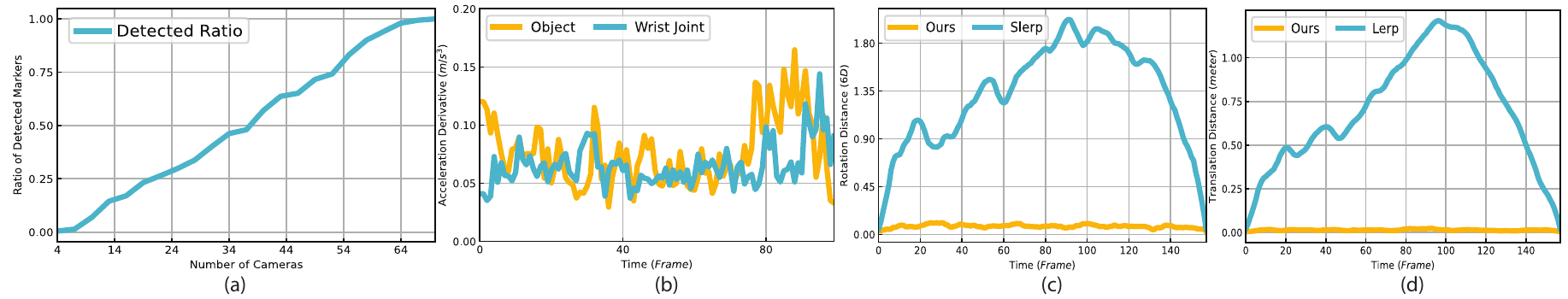}
    \caption{(a) Ratio of markers being detected with respect to 70 cameras to sampled number of cameras (b) Acceleration derivative of object and interacting hand wrist joint (c)  Comparison of difference between original translation and hole-filled versions (d) Comparison of difference between original rotation (in 6D) and hole-filled versions}
    \label{fig:post_proc}
    \vspace{-10pt}
\end{figure}

\noindent \textbf{Impact of Camera Number.}
We demonstrate the advantage of using all 70 cameras against the alternative solutions of using fewer cameras. To quantify the impact of camera number, we count the number of detected marker corners by simulating systems with varying numbers of cameras where the subsets are derived from the complete set of cameras.
As seen in Fig.~\ref{fig:post_proc}, the detected ratio tends to increase linearly with the addition of cameras to the subsets without saturation. This observation implies that we take full advantage of our system without experiencing redundancy. 
\begin{table}[t!]
    \centering
    \begin{subtable}{0.8\columnwidth}
        \centering
        \resizebox{\columnwidth}{!}{
            \begin{tabular}{l|c|c}
                \toprule
                & \textbf{Average Tracked} & \textbf{During Manipulation} \\
                \midrule
                Mean Cam. Num. (Visibility) & 8.94 & 8.59 \\
                Reproj. Error [pixel] & 1.021 & 1.016 \\
                \bottomrule
            \end{tabular}
        }
        \phantomsubcaption
        \label{tab:reproj}
    \end{subtable}
    \hfill
    \begin{subtable}{0.8\columnwidth}
        \centering
        \resizebox{\columnwidth}{!}{
            \begin{tabular}{l| c| c| c| c| c| c}
                \toprule
                IR Marker Num.& \textbf{Ours} & \textbf{4} & \textbf{7} & \textbf{10} & \textbf{20} & \textbf{40} \\ 
                \midrule
                Tracking Success Ratio & 1.0 & 0.76 & 0.79 & 0.86 & 0.90 & 0.93\\
                \bottomrule
            \end{tabular}
        }
        \phantomsubcaption
        \label{tab:virtual}
    \end{subtable}
    \caption{Evaluation on system settings (Up) Average Reprojection error detected in the scene and during manipulation by humans. (Down) Average number of tracked object ratio on multiple sampled windows. Numbers in the upper row represent number of virtual passive markers attached to the surface of the target object.}
    \label{tab:eval}
 \vspace{-10pt}
\end{table}

\noindent \textbf{Advantage of 3D Marker Cube.} %
We compare our 3D cube marker solution against an alternative solution of attaching markers on the surface of the targets %
(e.g., IR markers as in ~\cite{fan2023arctic}). To simulate such IR marker system, we sample a set of points on the object mesh surface, assuming the sampled points as virtual markers. We choose a 4-minute long sequence, where our 3D marker-based object tracking is fully successful. Then, we assess the tracking ratio of virtual surface markers, considering varying numbers of markers attached as shown in Tab.~\ref{tab:eval}. We simulate an occulsion to markers using human mesh, and assume tracking fails if they are visible from less than 3 cameras. Setting window length as 300, an object tracking fails %
if there exists any frame that less than 4 attached markers are tracked. Finally, we compute the average tracking success ratio in each marker setup, as shown in Tab.~\ref{tab:eval}. As seen, attaching markers on the object surface suffers from tracking failures mainly because the actors tend to touch the object during interactions and %
the existence of multiple objects in the scene.

\noindent \textbf{3D Marker Triangulation Quality.} To measure the quality of tracking objects via ArUco markers, we report the average reprojection error of triangulated ArUco corners for all tracked scenes and for the interaction target parts. 
We also report the average number of cameras (visibility) used in corner triangulation for each case, when in contact and averaged in all scenes. Check Tab.~\ref{tab:eval} for the details.%

\noindent \textbf{Tracking Quality via Assessing Temporal Jitters.}
We compare the quality of tracking of our two heterogeneous systems, a multi-view system and the IMU-based wearable mocap suits. As a way to verify the quality, we assess the temporal motion jitters of both system, by plotting the derivative of acceleration for a certain interval of time. For this test, we capture a dedicated sequence, interacting with small and movable objects such as a cup and a kettle by moving them into various poses in multiple areas. The result is shown in Fig.~\ref{fig:post_proc}, where both systems show similar jitter levels. Given that we use the commercial expert-level wearable motion capture system equipped with high-quality IMUs, this particular result demonstrates that our object tracking quality from cameras is comparable to the wearable capture system, enabling subtle interaction captures.

\noindent \textbf{Hand Alignment Evaluation.}
To quantify the quality of our hand alignment procedure shown in Sec.~\ref{sec:Capturing_3D_Human_Motions}, we perform a validation capture, where a participant touches random marker corners attached to the objects in random places inside the studio. Then, we measure the average distance between the fingertip and the target marker corner at the contact.
The Average Position Error (APE) is 11 $mm$ with 86 finger touches with various fingertips. Considering a possible bias %
due to the finger width(15-20 $mm$), our result demonstrates high precision in spatial alignment quality.

\subsection{Post Processing Validation}
We validate the performance of our post-processing method. As a way to quantify the performance, we simulate the tracking failure cases by intentionally dropping certain intervals of frames from successfully tracked sequences, and applying our post-processing method to recover the dropped frames. By comparing the originally tracked cues with the recovered cues from the post-processing, we can assess the quality of our post-processing outputs. 
 
\noindent \textbf{Evaluating Object Tracking Enhancement.}
We perform our object-tracking enhancement process using the cues from hand tracking and compare its quality by a naive interpolation method based on linear functions (e.g. \emph{lerp} for translation and \emph{slerp} for rotation). 
The quantitative results are shown in Fig.~\ref{fig:post_proc}, where our method powered with mocap device observations demonstrates much better performance in recovering the tracking failures. 

\noindent \textbf{Evaluating Hand Tracking Enhancement.}
To test the robustness of our system, we recover undetected markers of the dropped frames with our postprocessing algorithm. 
The error between hand joints computed using recovered marker positions and joint positions of original output is 9 $mm$.

 \section{Modeling Parameterized Episodic HOI}\label{formulation}
To learn the characteristic correlations between the environment $\mathbf{S}_e(t)$ changes and the human actions $\mathbf{S}_p(t)$, parameterizing both in a common spatio-temporal space is crucial, which motivates us to build our \ParaHouse system. 
One direction towards generative modeling for human-object interaction is to probabilistically model the distributions of possible configurations of humans and objects in the parametric 3D space: $P( \mathbf{S}(0:t) )$, where $\mathbf{S}(t) = \{ \mathbf{S}_e(t), \mathbf{S}_p(t) \}$ and $t$ being a sequence length from $0$ to $T$. Intuitively, this formulation captures the likelihood of plausible human and object configurations and dynamics.
We can alternatively formulate it as a prediction problem: $\mathbf{S}_o(0{\,:\,}T) =\mathcal{F}\left( \mathbf{S}_i(0{\,:\,}T) \right)$,
where the input $\mathbf{S}_i(0{\,:\,}T)$ is a subset of full states $\mathbf{S}(0{\,:\,}T)$ and $\mathbf{S}_o(0{\,:\,}T)$ being the reconstructed unseen cues predicted by the model. For instance, we can build the model to infer 3D object movements from the human body motions, finger motions from desired object movements or formulate the HOI understanding as a future status forecasting task: $\mathbf{S}_o(t+1{\,:\,}t+w) =\mathcal{F}\left( \mathbf{S}_i(t-w{\,:\,}t)\right)$.
Furthermore, such formulation can also include the text descriptions $\mathcal{T}$ as an additional input condition. All these formulations can be interesting future research directions as a way to understand and learn spatio-temporal relations of human-object interactions in casual and natural activities, to which our \ParaHouse system and collected dataset can contribute. This paper explores two possible example problems: text-conditioned motion synthesis and object-guided motion synthesis.

\begin{figure}[t!]
    \centering
    \includegraphics[width=\linewidth, trim={0cm 0.0cm 0cm 0cm},clip]{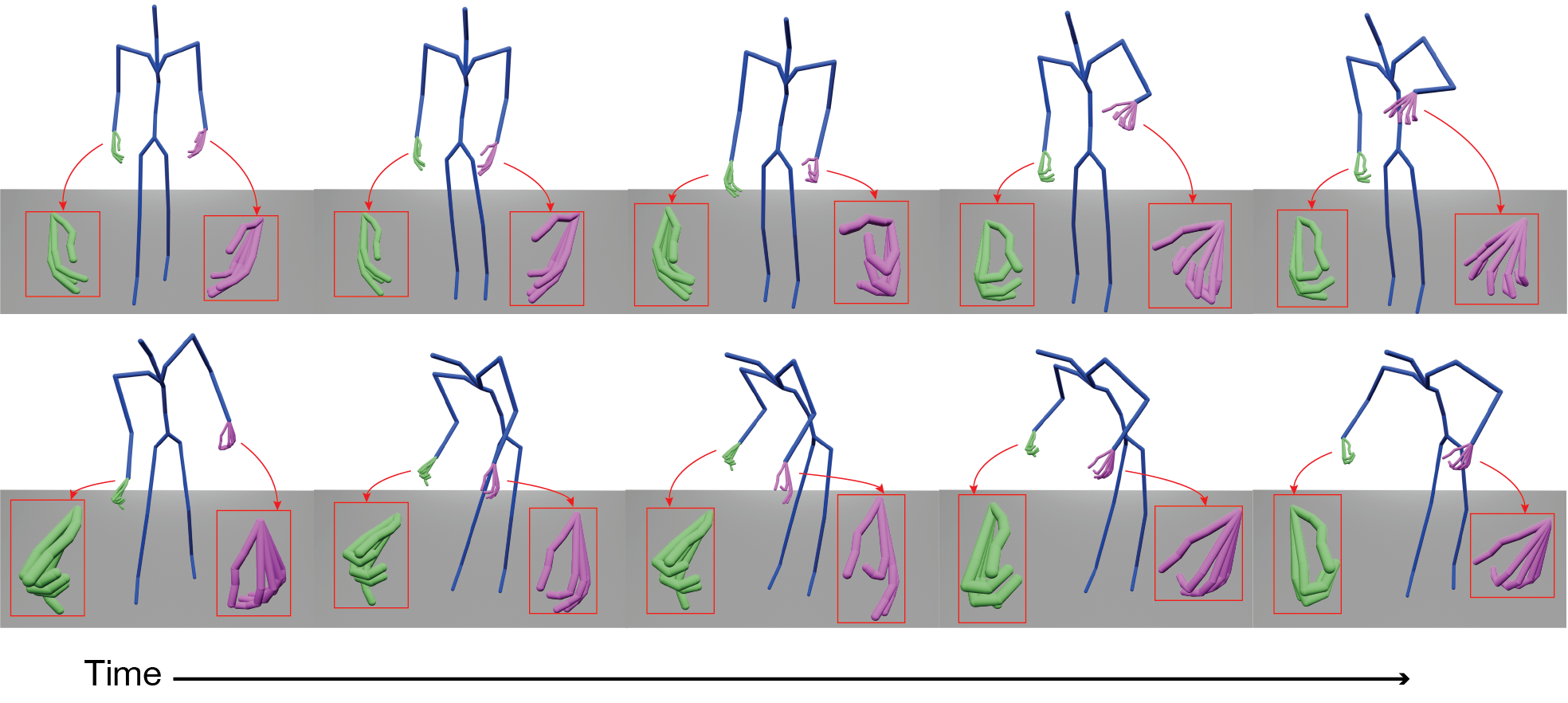}
    \caption{A sampled example of (Up) `A person walks shortly then lift kettle with left hand and pours toward cup holding right' (Down) `A person opens the cabinet'
    }
    \label{fig:sampled_skel}
    \vspace{-5pt}
\end{figure}

\subsection{Text Conditioned Motion Synthesis}
We consider a task of generating human motions both body and hands given text descriptions. In such context, we %
aim to train a model that covers both diverse human actions (e.g. running, jumping) and sophisticated finger-level manipulations(e.g. turning on gasstove, cutting ingredients). 
Since our dataset mainly focuses on capturing manipulations, we utilize current large-scale text-motion paired dataset HumanML3D ~\cite{Guo_2022_CVPR} to accomodate a diverse range of human motions.  Specifically, we downsample HumanML3D dataset to match the size of our dataset to balance the occurrence during training. 
For evaluation, we compare the models in three variations : model trained with ParaHome soley, model trained with HumanML3D soley and the model trained with evenly mixed dataset to see which model closely approximates the distribution of the entire combined dataset. We also qualitatively demonstrate that generating an intermediate action that contains both features from ParaHome and HumanML3D %
is possible.

\noindent \textbf{Formulation:} 
We define our problem as learning a data distribution using a conditional variational model $\mathcal{F}$, which generates a sequence $\mathbf{S}_p(t{\,:\,}t+w)$ from noise $z \sim \mathcal{N}(0, \mathbf{I})$ and condition $c$
, by training the model to reconstruct the data itself. Thus $\mathbf{S}_p(t{\,:\,}t+w) = \mathcal{F}(z, c) $ where $z \sim \mathcal{N}(0, \mathbf{I})$ and $c$ being text description.
For each time $t$, we define $\mathbf{S}_p(t)$ as a concatenation of root height $r_y \in \mathbb{R}^1$, root velocity $r_v \in \mathbb{R}^2$, root rotation $r_w \in \mathbb{R}^1$, localized joint position $j_p \in \mathbb{R}^{61\times3}$, joint relative rotation $j_w \in \mathbb{R}^{61\times6}$, joint local velocity $j_v \in \mathbb{R}^{61\times3}$, and foot contact $t_f \in \mathbb{R}^4$, thus $\mathbf{S}_p(t) = \{ r_y, r_v, r_w, j_p, j_w, j_v, t_f \} \in \mathbb{R}^{743}$ following %
 Text2Motion~\cite{Guo_2022_CVPR}. We use MDM~\cite{tevet2022human} framework which utilizes diffusion model $p_{\theta}$ to generate samples from noise $x_N \sim \mathcal{N}(0, \mathbf{I})$ via reverse process $$p_{\theta}(x_{n-1}|x_n, c) = \mathcal{N}(x_{n-1}; \mu_{\theta}(x_n, c), \sigma_n^2\mathbf{I})$$ where $x_n = \mathbf{S}_p^n$. Also instead of training model $\mu_{\theta}$ to estimate noise, we directly train model to get cleaned sample $\mathbf{S}_p^0$. Then optimization target in training the model is 
$$\mathcal{L} = \mathbb{E}_{ \mathbf{S}_p^0, n \sim [1,N]}\left\| \mathbf{S}_p^0 - \mathcal{F}(\mathbf{S}_p^n, n, c) \right\|_2^2$$  
 After training, generating samples given text condition $c$, we can invert the samples using DDIM and acquire an approximate latent vector~\cite{song2020denoising}. In order to generate an intermediate action between generated samples $S^0_{p,1}$ from ParaHome and $S^0_{p,2}$ from HumanML3D, we invert the samples into 
 $\tilde{S}^N_{p,1}$ and $\tilde{S}^N_{p,2}$ via the above process.
 Then we interpolate the latents $\tilde{S}^N_{p,\text{interp}} = u\tilde{S}^N_{p,1} + (1-u)\tilde{S}^N_{p,2}$ and run reverse diffusion process to get the motion $ S^0_{p,\text{interp}}$. 

\begin{figure}[t!]
    \centering
    \includegraphics[width=\linewidth, trim={0cm 0.0cm 0cm 0cm},clip]{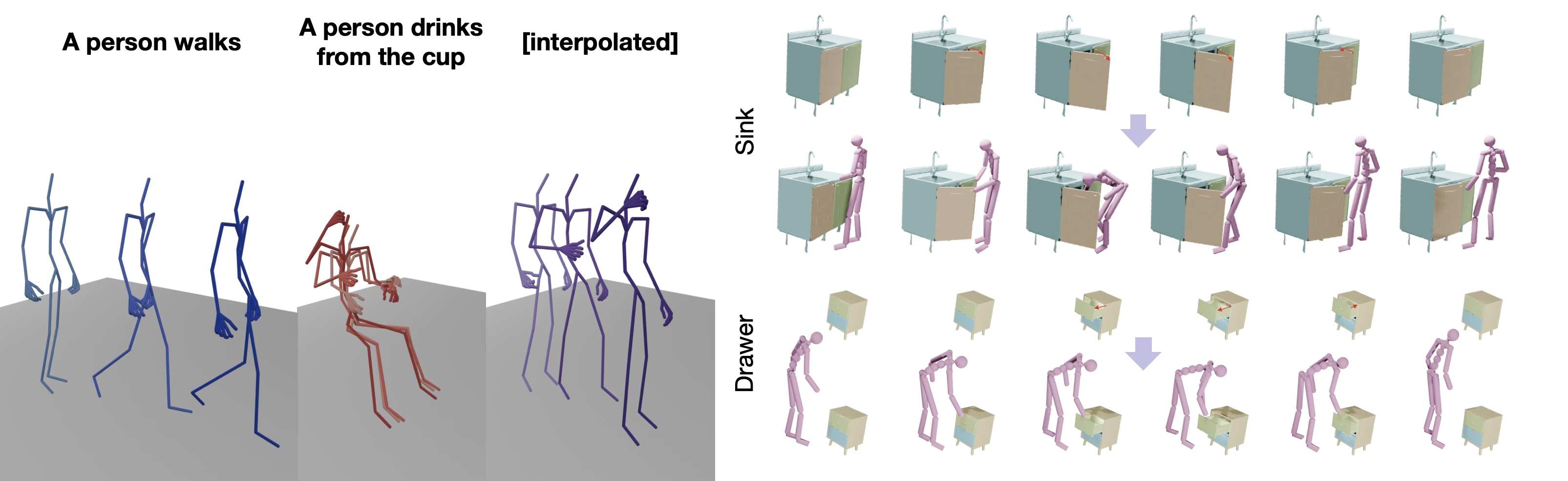}
    \caption{(Left) A sampled example of latent interpolation result. Right most is a sequence generated from the interpolated noise. (Right) Generated body motion(lower) given object motion trajectory from 0 to T(upper).
    }
    \label{fig:interp}
\end{figure}

\noindent \textbf{Results:}  
Following MDM, we measure $\textit{R-Precision}$, $\textit{Multimodal Distance}$, $\textit{FID}$, $\textit{Diversity}$, and $\textit{Multimodality}$ to compare each trained model's expressiveness and distance to the distribution of unseen test set of the mixed dataset. As in Tab.~\ref{tab:dataset_comp}, the results on $\textit{R-Precision}$, $\textit{FID}$ and $\textit{Diversity}$ show that the training with evenly mixed data describes both dataset without degrading both motions. This proves that our dataset and HumanML3D can be well blended and qualitative samples are shown in Fig.~\ref{fig:sampled_skel}. Also, we qualitatively demonstrate a denoised result of the interpolated latent noise, where the original latents are inverted from samples generated with text descriptions from each dataset.
The model trained with evenly mixed dataset and current algorithm have capability of mixing both distinct styles of data (i.e. variety of human movement motions in HumanML3D and manipulating actions in our dataset) thus can be used in generating intermediate actions as in Fig.~\ref{fig:interp} (Left) where a person walks and drinks simultaneously.

\begin{table}[tb]
\centering
\resizebox{\columnwidth}{!}{
\begin{tabular}{lcccccc}
\toprule
\textbf{Method} & \textbf{R Precision (top3)}↑ & \textbf{Mutimodal. dist}↓ & \textbf{FID}↓ &  \textbf{Diversity}→ & \textbf{Multimodality}↑ \\
\midrule
Real & $ 0.84 $ & $ 2.75 $ & $ 0.07  $ & $ 15.50 $ & $ - $ \\ 
\midrule
ParaHome & $ 0.40  $ & $ 12.46 $ & $ 148.31 $ & $ 10.13 $ & $\boldsymbol{ 3.52 }$ \\
Humanml3D & $ 0.46 $ & $ 8.26 $ & $ 74.66 $ & $ 8.21 $ & $ 2.23 $ \\
Mixed & $ \boldsymbol{0.73 } $ & $\boldsymbol{ 3.75 }$ & $\boldsymbol{ 0.85 }$ & $\boldsymbol{ 15.06 }$ & $ 2.37 $ \\
\bottomrule
\end{tabular}
}
\caption{Evaluation of models trained with different set of data}
\label{tab:dataset_comp}
 \vspace{-10pt}
\end{table}

\subsection{Object-guided Body Motion Synthesis}\label{noveltask}

In this section, we consider the scenario of synthesizing a sequence of full-body motions given a sequence of object states. This is to investigate the viability of learning a spatio-temporal correlation between human motions and object state changes.

\noindent \textbf{Formulation:} 
We formulate the problem as, $\mathbf{S}_p(t{\,:\,}t+w) =\mathcal{F}_{o\rightarrow p}\left( \mathbf{S}_{to}(t{\,:\,}t+w), S_{p \rightarrow o}(t) \right)$, where $\mathbf{S}_{to}(t{\,:\,}t+w) \in \mathbf{S}_{e}(t{\,:\,}t+w)$ is the desired status changes of a target object including rigid body motion and object-specific articulated parts representation. $S_{p \rightarrow o}(t)$ is the relative location and orientation of the person's root towards the object at the initial time. The output  $\mathbf{S}_p(t{\,:\,}t+w)$ is the synthesized human motion in the corresponding time window. For representing human motion, we use both person root-centric coordinates as in ~\cite{holden2016deep} and global coordinates. As a model for achieving this task, we extend a transformer-based model for text to 3D human motion synthesis~\cite{petrovich2022temos}, by introducing Object State Encoder $\mathcal{O}_{enc}$ as a replacement of the text encoder module. We provide more details on data representation, model architecture and experimental results for alternative representation options in our supp. mat..

\noindent \textbf{Results}: %
To assess the quality of generated human body motions, 
we quantitatively measure mean positional errors from the ground truth and the output in two different coordinate systems : (1) root-centered coordinate, (2) global coordinate. For case (1), we compute entire joints error (\emph{rc-joints}) and wrist error (\emph{rc-wrist}). For case (2), we compute root-position error (\emph{glb-root}) and global-joints error (\emph{glb-joints}). Additionally we compute foot-skating as in \cite{he2022nemf}. The quantitative results are shown in Tab.~\ref{tab:quantitative_benchmark1} including qualitative results in Fig~\ref{fig:interp}. %
The qualitative results demonstrate convincing performance in synthesizing plausible human motions for the target tasks. Also, using relative spatial cues as input (\emph{with $S_{p \rightarrow tp}(0)$}) contributes to more accurate motion synthesis. We add further analysis regarding the results in supp. mat..

\begin{table}[tb]
  \centering
  \resizebox{\columnwidth}{!}{
    \begin{tabular}{c|c|cccc|c|c}
      \toprule
      \multicolumn{1}{c}{} & \multicolumn{1}{c}{} &  \multicolumn{4}{c}{\textbf{MPE$\downarrow$}(cm)} & \multicolumn{1}{c}{} & \multicolumn{1}{c}{}\\
      \cmidrule(rl){3-6} 
      Object & condition & rc-joints & rc-wrists & glb-root & glb-joints & \textbf{MOE$\downarrow$} &
      \textbf{Foot skating}$\downarrow$\\
      \midrule
       \multirow{2}{*}{Refrigerator} 
       & w/o $S_{p \rightarrow tp}$ & 
       1.05 & 2.66 & 7.79 & 9.26 & 0.85& \textbf{1.43} \\
       &  w/ $S_{p \rightarrow tp}$ & \textbf{0.97} & \textbf{2.09} & \textbf{5.60} & \textbf{6.57} & \textbf{0.75}& 1.74\\
      \midrule
      \multirow{2}{*}{Drawer} 
       & w/o $S_{p \rightarrow tp}$ & 3.81 & 5.25 & 5.58 & 9.92 & 0.66 & \textbf{1.01}\\
       &  w/ $S_{p \rightarrow tp}$ & \textbf{2.61}& \textbf{2.90} & \textbf{3.92} & \textbf{5.88} & 0.66 & 1.58 \\
      \midrule
      \multirow{2}{*}{Sink} 
       & w/o $S_{p \rightarrow tp}$ & 
                                      1.32 & 2.86 & 6.94 & 8.32 & 0.79 & \textbf{1.08}\\
       &  w/ $S_{p \rightarrow tp}$ & \textbf{1.21} & \textbf{1.93} & \textbf{3.83} & \textbf{4.84} & \textbf{0.76} & 1.67 \\
      \midrule
      \multirow{2}{*}{\makecell{Washing\\machine}} 
       & w/o $S_{p \rightarrow tp}$ & 
                                      1.33 & 3.14 & 11.29 & 14.43 & 1.00 & \textbf{2.00}\\
       &  w/ $S_{p \rightarrow tp}$ & \textbf{1.29} & \textbf{2.62} & \textbf{9.03} & \textbf{11.46} & \textbf{0.92} & 2.60  \\
    \bottomrule
    \end{tabular}
  }
  \caption{Quantitative results of object guided motion synthesis task.}
  \label{tab:quantitative_benchmark1}
  \vspace{-10pt}
\end{table}

\section{Discussion}
We present a \ParaHouse system, specifically designed to capture human motions, finger movements, and object dynamics in 3D during natural human-object interactions in a studio apartment setting. 
Leveraging our system, we collect a new HOI dataset, which offers key improvements over existing datasets, including (1) capturing 3D body and hand motion alongside 3D object movement within a contextual home environment; (2) encompassing human interaction with multiple objects and concurrent usage of objects; (3) including articulated objects with multiple parts and diverse scale objects. We explore %
the opportunities in leveraging our dataset for human motion synthesis via generative modeling.

We endeavor to improve our system further to handle limitations. Firstly, we plan to evolve our system towards a markerless approach by eliminating artificial markers and constructing motion priors from the current database. Secondly, we aim to replicate our system in different room settings, with more diverse objects ensuring better generalizability.

{
    \small
    \bibliographystyle{ieeenat_fullname}
    \bibliography{11_references}

\begin{thebibliography}{130}
\providecommand{\natexlab}[1]{#1}
\providecommand{\url}[1]{\texttt{#1}}
\expandafter\ifx\csname urlstyle\endcsname\relax
  \providecommand{\doi}[1]{doi: #1}\else
  \providecommand{\doi}{doi: \begingroup \urlstyle{rm}\Url}\fi

\bibitem[Aristidou et~al.(2016)Aristidou, Chrysanthou, and Lasenby]{Aristidou:2016:ExtFABRIK}
Andreas Aristidou, Yiorgos Chrysanthou, and Joan Lasenby.
\newblock Extending {FABRIK} with model constraints.
\newblock \emph{Comput. Animat. Virtual Worlds}, 2016.

\bibitem[Bhatnagar et~al.(2022)Bhatnagar, Xie, Petrov, Sminchisescu, Theobalt, and Pons-Moll]{bhatnagar2022behave}
Bharat~Lal Bhatnagar, Xianghui Xie, Ilya~A Petrov, Cristian Sminchisescu, Christian Theobalt, and Gerard Pons-Moll.
\newblock Behave: Dataset and method for tracking human object interactions.
\newblock \emph{CVPR}, 2022.

\bibitem[{Black Forest Lab}(2024)]{flux}
{Black Forest Lab}, 2024.
\newblock \url{https://huggingface.co/black-forest-labs}.

\bibitem[Brahmbhatt et~al.(2020)Brahmbhatt, Tang, Twigg, Kemp, and Hays]{brahmbhatt2020contactpose}
Samarth Brahmbhatt, Chengcheng Tang, Christopher~D Twigg, Charles~C Kemp, and James Hays.
\newblock Contactpose: A dataset of grasps with object contact and hand pose.
\newblock \emph{ECCV}, 2020.

\bibitem[Bregler et~al.(2004)Bregler, Malik, and Pullen]{Bregler-04}
Christoph Bregler, Jitendra Malik, and Katherine Pullen.
\newblock {Twist based acquisition and tracking of animal and human kinematics}.
\newblock \emph{IJCV}, 2004.

\bibitem[Brox et~al.(2010)Brox, Rosenhahn, Gall, and Cremers]{Brox-10}
Thomas Brox, Bodo Rosenhahn, Juergen Gall, and Daniel Cremers.
\newblock {Combined region and motion-based 3D tracking of rigid and articulated objects.}
\newblock \emph{TPAMI}, 2010.

\bibitem[Cao et~al.(2017)Cao, Simon, Wei, and Sheikh]{cao2017realtime}
Zhe Cao, Tomas Simon, Shih-En Wei, and Yaser Sheikh.
\newblock Realtime multi-person 2d pose estimation using part affinity fields.
\newblock 2017.

\bibitem[Cao et~al.(2020)Cao, Gao, Mangalam, Cai, Vo, and Malik]{cao2020long}
Zhe Cao, Hang Gao, Karttikeya Mangalam, Qi-Zhi Cai, Minh Vo, and Jitendra Malik.
\newblock Long-term human motion prediction with scene context.
\newblock \emph{ECCV}, 2020.

\bibitem[Cao et~al.(2021)Cao, Radosavovic, Kanazawa, and Malik]{cao2021reconstructing}
Zhe Cao, Ilija Radosavovic, Angjoo Kanazawa, and Jitendra Malik.
\newblock Reconstructing hand-object interactions in the wild.
\newblock In \emph{ICCV}, 2021.

\bibitem[Cha et~al.(2024)Cha, Kim, Yoon, and Baek]{cha2024text2hoi}
Junuk Cha, Jihyeon Kim, Jae~Shin Yoon, and Seungryul Baek.
\newblock Text2hoi: Text-guided 3d motion generation for hand-object interaction.
\newblock pages 1577--1585, 2024.

\bibitem[Chao et~al.(2021)Chao, Yang, Xiang, Molchanov, Handa, Tremblay, Narang, Van~Wyk, Iqbal, Birchfield, et~al.]{chao2021dexycb}
Yu-Wei Chao, Wei Yang, Yu Xiang, Pavlo Molchanov, Ankur Handa, Jonathan Tremblay, Yashraj~S Narang, Karl Van~Wyk, Umar Iqbal, Stan Birchfield, et~al.
\newblock Dexycb: A benchmark for capturing hand grasping of objects.
\newblock \emph{CVPR}, 2021.

\bibitem[Cheung et~al.(2005)Cheung, Baker, and Kanade]{Cheung-05}
Kong~Man Cheung, Simon Baker, and Takeo Kanade.
\newblock {Shape-from-silhouette across time part i: Theory and algorithms}.
\newblock \emph{IJCV}, 2005.

\bibitem[Christen et~al.(2022)Christen, Kocabas, Aksan, Hwangbo, Song, and Hilliges]{christen2022d}
Sammy Christen, Muhammed Kocabas, Emre Aksan, Jemin Hwangbo, Jie Song, and Otmar Hilliges.
\newblock D-grasp: Physically plausible dynamic grasp synthesis for hand-object interactions.
\newblock pages 20577--20586, 2022.

\bibitem[Christen et~al.(2024)Christen, Hampali, Sener, Remelli, Hodan, Sauser, Ma, and Tekin]{christen2024diffh2o}
Sammy Christen, Shreyas Hampali, Fadime Sener, Edoardo Remelli, Tomas Hodan, Eric Sauser, Shugao Ma, and Bugra Tekin.
\newblock Diffh2o: Diffusion-based synthesis of hand-object interactions from textual descriptions.
\newblock \emph{arXiv preprint arXiv:2403.17827}, 2024.

\bibitem[Corazza et~al.(2010)Corazza, M{\"{u}}ndermann, Gambaretto, Ferrigno, and Andriacchi]{Corazza-10}
Stefano Corazza, Lars M{\"{u}}ndermann, Emiliano Gambaretto, Giancarlo Ferrigno, and Thomas~P. Andriacchi.
\newblock {Markerless Motion Capture through Visual Hull, Articulated ICP and Subject Specific Model Generation}.
\newblock \emph{IJCV}, 2010.

\bibitem[Corona et~al.(2020)Corona, Pumarola, Alenya, Moreno-Noguer, and Rogez]{corona2020ganhand}
Enric Corona, Albert Pumarola, Guillem Alenya, Francesc Moreno-Noguer, and Gr{\'e}gory Rogez.
\newblock Ganhand: Predicting human grasp affordances in multi-object scenes.
\newblock \emph{CVPR}, 2020.

\bibitem[Dai et~al.(2022)Dai, Lin, Wen, Shen, Xu, Yu, Ma, and Wang]{dai2022hsc4d}
Yudi Dai, Yitai Lin, Chenglu Wen, Siqi Shen, Lan Xu, Jingyi Yu, Yuexin Ma, and Cheng Wang.
\newblock Hsc4d: Human-centered 4d scene capture in large-scale indoor-outdoor space using wearable imus and lidar.
\newblock 2022.

\bibitem[de~Aguiar et~al.(2008)de~Aguiar, Stoll, Theobalt, Ahmed, Seidel, and Thrun]{deAguiar-2008}
Edilson de Aguiar, Carsten Stoll, Christian Theobalt, Naveed Ahmed, Hans-Peter Seidel, and Sebastian Thrun.
\newblock Performance capture from sparse multi-view video.
\newblock \emph{SIGGRAPH}, 2008.

\bibitem[DelPreto et~al.(2022)DelPreto, Liu, Luo, Foshey, Li, Torralba, Matusik, and Rus]{delpreto2022actionsense}
Joseph DelPreto, Chao Liu, Yiyue Luo, Michael Foshey, Yunzhu Li, Antonio Torralba, Wojciech Matusik, and Daniela Rus.
\newblock Actionsense: A multimodal dataset and recording framework for human activities using wearable sensors in a kitchen environment.
\newblock \emph{NeurIPS}, 35:\penalty0 13800--13813, 2022.

\bibitem[Diller and Dai(2023)]{diller2023cg}
Christian Diller and Angela Dai.
\newblock Cg-hoi: Contact-guided 3d human-object interaction generation.
\newblock \emph{arXiv preprint arXiv:2311.16097}, 2023.

\bibitem[Doosti et~al.(2020)Doosti, Naha, Mirbagheri, and Crandall]{doosti2020hope}
Bardia Doosti, Shujon Naha, Majid Mirbagheri, and David~J Crandall.
\newblock Hope-net: A graph-based model for hand-object pose estimation.
\newblock In \emph{CVPR}, 2020.

\bibitem[{Einstar}(2023)]{einstar}
{Einstar}, 2023.
\newblock \url{https://www.einstar.com/}.

\bibitem[Fan et~al.(2023)Fan, Taheri, Tzionas, Kocabas, Kaufmann, Black, and Hilliges]{fan2023arctic}
Zicong Fan, Omid Taheri, Dimitrios Tzionas, Muhammed Kocabas, Manuel Kaufmann, Michael~J Black, and Otmar Hilliges.
\newblock Arctic: A dataset for dexterous bimanual hand-object manipulation.
\newblock 2023.

\bibitem[Furukawa and Ponce(2008)]{Furukawa-2008}
Y. Furukawa and J. Ponce.
\newblock Dense 3d motion capture from synchronized video streams.
\newblock \emph{CVPR}, 2008.

\bibitem[Gall et~al.(2009)Gall, Stoll, De~Aguiar, Theobalt, Rosenhahn, and Seidel]{Gall-09}
Juergen Gall, Carsten Stoll, Edilson De~Aguiar, Christian Theobalt, Bodo Rosenhahn, and Hans-Peter Seidel.
\newblock Motion capture using joint skeleton tracking and surface estimation.
\newblock \emph{CVPR}, 2009.

\bibitem[Garcia-Hernando et~al.(2018)Garcia-Hernando, Yuan, Baek, and Kim]{garcia2018first}
Guillermo Garcia-Hernando, Shanxin Yuan, Seungryul Baek, and Tae-Kyun Kim.
\newblock First-person hand action benchmark with rgb-d videos and 3d hand pose annotations.
\newblock \emph{CVPR}, 2018.

\bibitem[Garrido-Jurado et~al.(2014)Garrido-Jurado, Mu{\~n}oz-Salinas, Madrid-Cuevas, and Mar{\'\i}n-Jim{\'e}nez]{garrido2014automatic}
Sergio Garrido-Jurado, Rafael Mu{\~n}oz-Salinas, Francisco~Jos{\'e} Madrid-Cuevas, and Manuel~Jes{\'u}s Mar{\'\i}n-Jim{\'e}nez.
\newblock Automatic generation and detection of highly reliable fiducial markers under occlusion.
\newblock \emph{Pattern Recognition}, 47\penalty0 (6):\penalty0 2280--2292, 2014.

\bibitem[Gavrila and Davis(1996)]{Gavrila-96}
D. Gavrila and LS Davis.
\newblock {Tracking of humans in action: A 3-D model-based approach}.
\newblock \emph{ARPA Image Understanding Workshop}, 1996.

\bibitem[Ghosh et~al.(2023)Ghosh, Dabral, Golyanik, Theobalt, and Slusallek]{ghosh2023imos}
Anindita Ghosh, Rishabh Dabral, Vladislav Golyanik, Christian Theobalt, and Philipp Slusallek.
\newblock Imos: Intent-driven full-body motion synthesis for human-object interactions.
\newblock In \emph{Computer Graphics Forum}, pages 1--12. Wiley Online Library, 2023.

\bibitem[Grady et~al.(2021)Grady, Tang, Twigg, Vo, Brahmbhatt, and Kemp]{grady2021contactopt}
Patrick Grady, Chengcheng Tang, Christopher~D Twigg, Minh Vo, Samarth Brahmbhatt, and Charles~C Kemp.
\newblock Contactopt: Optimizing contact to improve grasps.
\newblock In \emph{CVPR}, 2021.

\bibitem[Gross et~al.(2003)Gross, W\"{u}rmlin, Naef, Lamboray, Spagno, Kunz, Koller-Meier, Svoboda, Van~Gool, Lang, Strehlke, Moere, and Staadt]{Gross-2003}
Markus Gross, Stephan W\"{u}rmlin, Martin Naef, Edouard Lamboray, Christian Spagno, Andreas Kunz, Esther Koller-Meier, Tomas Svoboda, Luc Van~Gool, Silke Lang, Kai Strehlke, Andrew~Vande Moere, and Oliver Staadt.
\newblock Blue-c: A spatially immersive display and 3d video portal for telepresence.
\newblock \emph{SIGGRAPH}, 2003.

\bibitem[Gross and Shi(2001)]{gross2001cmu}
Ralph Gross and Jianbo Shi.
\newblock The cmu motion of body (mobo) database.
\newblock 2001.

\bibitem[Guo et~al.(2022{\natexlab{a}})Guo, Zou, Zuo, Wang, Ji, Li, and Cheng]{Guo_2022_CVPR}
Chuan Guo, Shihao Zou, Xinxin Zuo, Sen Wang, Wei Ji, Xingyu Li, and Li Cheng.
\newblock Generating diverse and natural 3d human motions from text.
\newblock In \emph{Proceedings of the IEEE/CVF Conference on Computer Vision and Pattern Recognition (CVPR)}, 2022{\natexlab{a}}.

\bibitem[Guo et~al.(2022{\natexlab{b}})Guo, Zou, Zuo, Wang, Ji, Li, and Cheng]{guo2022generating}
Chuan Guo, Shihao Zou, Xinxin Zuo, Sen Wang, Wei Ji, Xingyu Li, and Li Cheng.
\newblock Generating diverse and natural 3d human motions from text.
\newblock \emph{CVPR}, 2022{\natexlab{b}}.

\bibitem[Guzov et~al.(2021)Guzov, Mir, Sattler, and Pons-Moll]{guzov2021human}
Vladimir Guzov, Aymen Mir, Torsten Sattler, and Gerard Pons-Moll.
\newblock Human poseitioning system (hps): 3d human pose estimation and self-localization in large scenes from body-mounted sensors.
\newblock \emph{CVPR}, 2021.

\bibitem[Hampali et~al.(2020)Hampali, Rad, Oberweger, and Lepetit]{hampali2020honnotate}
Shreyas Hampali, Mahdi Rad, Markus Oberweger, and Vincent Lepetit.
\newblock Honnotate: A method for 3d annotation of hand and object poses.
\newblock \emph{CVPR}, 2020.

\bibitem[Hampali et~al.(2022)Hampali, Sarkar, Rad, and Lepetit]{hampali2022keypoint}
Shreyas Hampali, Sayan~Deb Sarkar, Mahdi Rad, and Vincent Lepetit.
\newblock Keypoint transformer: Solving joint identification in challenging hands and object interactions for accurate 3d pose estimation.
\newblock \emph{CVPR}, 2022.

\bibitem[Han et~al.(2018)Han, Liu, Wang, Ye, Twigg, and Kin]{han2018online}
Shangchen Han, Beibei Liu, Robert Wang, Yuting Ye, Christopher~D Twigg, and Kenrick Kin.
\newblock Online optical marker-based hand tracking with deep labels.
\newblock \emph{ACM TOG}, 37\penalty0 (4):\penalty0 1--10, 2018.

\bibitem[Hassan et~al.(2019)Hassan, Choutas, Tzionas, and Black]{hassan2019resolving}
Mohamed Hassan, Vasileios Choutas, Dimitrios Tzionas, and Michael~J Black.
\newblock Resolving 3d human pose ambiguities with 3d scene constraints.
\newblock In \emph{ICCV}, 2019.

\bibitem[Hassan et~al.(2021)Hassan, Ceylan, Villegas, Saito, Yang, Zhou, and Black]{hassan2021samp}
Mohamed Hassan, Duygu Ceylan, Ruben Villegas, Jun Saito, Jimei Yang, Yi Zhou, and Michael Black.
\newblock Stochastic scene-aware motion prediction.
\newblock In \emph{ICCV}, 2021.

\bibitem[Hassan et~al.(2023)Hassan, Guo, Wang, Black, Fidler, and Peng]{hassan2023synthesizing}
Mohamed Hassan, Yunrong Guo, Tingwu Wang, Michael Black, Sanja Fidler, and Xue~Bin Peng.
\newblock Synthesizing physical character-scene interactions.
\newblock \emph{arXiv preprint arXiv:2302.00883}, 2023.

\bibitem[Hasson et~al.(2019{\natexlab{a}})Hasson, Varol, Tzionas, Kalevatykh, Black, Laptev, and Schmid]{hasson19_obman}
Yana Hasson, G{\"u}l Varol, Dimitrios Tzionas, Igor Kalevatykh, Michael~J. Black, Ivan Laptev, and Cordelia Schmid.
\newblock Learning joint reconstruction of hands and manipulated objects.
\newblock In \emph{CVPR}, 2019{\natexlab{a}}.

\bibitem[Hasson et~al.(2019{\natexlab{b}})Hasson, Varol, Tzionas, Kalevatykh, Black, Laptev, and Schmid]{hasson2019learning}
Yana Hasson, Gul Varol, Dimitrios Tzionas, Igor Kalevatykh, Michael~J Black, Ivan Laptev, and Cordelia Schmid.
\newblock Learning joint reconstruction of hands and manipulated objects.
\newblock In \emph{CVPR}, 2019{\natexlab{b}}.

\bibitem[He et~al.(2022)He, Saito, Zachary, Rushmeier, and Zhou]{he2022nemf}
Chengan He, Jun Saito, James Zachary, Holly Rushmeier, and Yi Zhou.
\newblock Nemf: Neural motion fields for kinematic animation.
\newblock \emph{NeurIPS}, 35:\penalty0 4244--4256, 2022.

\bibitem[Holden et~al.(2016)Holden, Saito, and Komura]{holden2016deep}
Daniel Holden, Jun Saito, and Taku Komura.
\newblock A deep learning framework for character motion synthesis and editing.
\newblock \emph{ACM Transactions on Graphics (TOG)}, 2016.

\bibitem[Huang et~al.(2022)Huang, Taheri, Black, and Tzionas]{huang2022intercap}
Yinghao Huang, Omid Taheri, Michael~J Black, and Dimitrios Tzionas.
\newblock Intercap: Joint markerless 3d tracking of humans and objects in interaction.
\newblock \emph{GCPR}, 2022.

\bibitem[Ionescu et~al.(2013)Ionescu, Papava, Olaru, and Sminchisescu]{ionescu2013human3}
Catalin Ionescu, Dragos Papava, Vlad Olaru, and Cristian Sminchisescu.
\newblock Human3. 6m: Large scale datasets and predictive methods for 3d human sensing in natural environments.
\newblock \emph{IEEE transactions on pattern analysis and machine intelligence}, 36\penalty0 (7):\penalty0 1325--1339, 2013.

\bibitem[Jian et~al.(2023)Jian, Liu, Li, Hu, and Liu]{jian2023affordpose}
Juntao Jian, Xiuping Liu, Manyi Li, Ruizhen Hu, and Jian Liu.
\newblock Affordpose: A large-scale dataset of hand-object interactions with affordance-driven hand pose.
\newblock pages 14713--14724, 2023.

\bibitem[Jiang et~al.(2023)Jiang, Liu, Cao, Cui, Zhang, Chen, Wang, Zhu, and Huang]{jiang2023full}
Nan Jiang, Tengyu Liu, Zhexuan Cao, Jieming Cui, Zhiyuan Zhang, Yixin Chen, He Wang, Yixin Zhu, and Siyuan Huang.
\newblock Full-body articulated human-object interaction.
\newblock In \emph{ICCV}, 2023.

\bibitem[Jiang et~al.(2024)Jiang, Zhang, Li, Ma, Wang, Chen, Liu, Zhu, and Huang]{jiang2024scaling}
Nan Jiang, Zhiyuan Zhang, Hongjie Li, Xiaoxuan Ma, Zan Wang, Yixin Chen, Tengyu Liu, Yixin Zhu, and Siyuan Huang.
\newblock Scaling up dynamic human-scene interaction modeling.
\newblock 2024.

\bibitem[Jome~Yazdian et~al.(2023)Jome~Yazdian, Liu, Cheng, and Lim]{jome2023motionscript}
Payam Jome~Yazdian, Eric Liu, Li Cheng, and Angelica Lim.
\newblock Motionscript: Natural language descriptions for expressive 3d human motions.
\newblock \emph{arXiv e-prints}, pages arXiv--2312, 2023.

\bibitem[Joo et~al.(2015)Joo, Liu, Tan, Gui, Nabbe, Matthews, Kanade, Nobuhara, and Sheikh]{joo2015panoptic}
Hanbyul Joo, Hao Liu, Lei Tan, Lin Gui, Bart Nabbe, Iain Matthews, Takeo Kanade, Shohei Nobuhara, and Yaser Sheikh.
\newblock Panoptic studio: A massively multiview system for social motion capture.
\newblock \emph{CVPR}, 2015.

\bibitem[Kanade et~al.(1997)Kanade, Rander, and Narayanan]{Kanade-1997}
Takeo Kanade, Peter Rander, and P.J. Narayanan.
\newblock Virtualized reality: Constructing virtual worlds from real scenes.
\newblock \emph{IEEE Multimedia}, 1997.

\bibitem[Kaufmann et~al.(2023)Kaufmann, Song, Guo, Shen, Jiang, Tang, Z{\'a}rate, and Hilliges]{kaufmann2023emdb}
Manuel Kaufmann, Jie Song, Chen Guo, Kaiyue Shen, Tianjian Jiang, Chengcheng Tang, Juan~Jos{\'e} Z{\'a}rate, and Otmar Hilliges.
\newblock Emdb: The electromagnetic database of global 3d human pose and shape in the wild.
\newblock In \emph{ICCV}, pages 14632--14643, 2023.

\bibitem[Ke et~al.(2023)Ke, Ye, Danelljan, Liu, Tai, Tang, and Yu]{sam_hq}
Lei Ke, Mingqiao Ye, Martin Danelljan, Yifan Liu, Yu-Wing Tai, Chi-Keung Tang, and Fisher Yu.
\newblock Segment anything in high quality.
\newblock \emph{arXiv:2306.01567}, 2023.

\bibitem[Kehl and Gool(2006)]{Kehl-06}
Roland Kehl and Luc~Van Gool.
\newblock {Markerless tracking of complex human motions from multiple views}.
\newblock \emph{CVIU}, 2006.

\bibitem[Kulkarni et~al.(2023)Kulkarni, Rempe, Genova, Kundu, Johnson, Fouhey, and Guibas]{kulkarni2023nifty}
Nilesh Kulkarni, Davis Rempe, Kyle Genova, Abhijit Kundu, Justin Johnson, David Fouhey, and Leonidas Guibas.
\newblock Nifty: Neural object interaction fields for guided human motion synthesis.
\newblock \emph{arXiv preprint arXiv:2307.07511}, 2023.

\bibitem[Kwon et~al.(2021)Kwon, Tekin, St{\"u}hmer, Bogo, and Pollefeys]{kwon2021h2o}
Taein Kwon, Bugra Tekin, Jan St{\"u}hmer, Federica Bogo, and Marc Pollefeys.
\newblock H2o: Two hands manipulating objects for first person interaction recognition.
\newblock \emph{ICCV}, 2021.

\bibitem[Lea et~al.(2017)Lea, Flynn, Vidal, Reiter, and Hager]{lea2017temporal}
Colin Lea, Michael~D Flynn, Rene Vidal, Austin Reiter, and Gregory~D Hager.
\newblock Temporal convolutional networks for action segmentation and detection.
\newblock 2017.

\bibitem[Lee and Joo(2024)]{lee2024mocap}
Jiye Lee and Hanbyul Joo.
\newblock Mocap everyone everywhere: Lightweight motion capture with smartwatches and a head-mounted camera.
\newblock \emph{CVPR}, 2024.

\bibitem[Li et~al.(2023)Li, Wu, and Liu]{li2023object}
Jiaman Li, Jiajun Wu, and C~Karen Liu.
\newblock Object motion guided human motion synthesis.
\newblock \emph{ACM Transactions on Graphics (TOG)}, 2023.

\bibitem[Li et~al.(2024)Li, Clegg, Mottaghi, Wu, Puig, and Liu]{li2023controllable}
Jiaman Li, Alexander Clegg, Roozbeh Mottaghi, Jiajun Wu, Xavier Puig, and C~Karen Liu.
\newblock Controllable human-object interaction synthesis.
\newblock \emph{ECCV}, 2024.

\bibitem[Liang et~al.(2023)Liang, He, Zhao, Li, Wang, Yu, and Xu]{liang2023hybridcap}
Han Liang, Yannan He, Chengfeng Zhao, Mutian Li, Jingya Wang, Jingyi Yu, and Lan Xu.
\newblock Hybridcap: Inertia-aid monocular capture of challenging human motions.
\newblock In \emph{AAAI}, 2023.

\bibitem[Liu et~al.(2023{\natexlab{a}})Liu, Chen, Ye, and Qi]{liu2023prior}
Jianhui Liu, Yukang Chen, Xiaoqing Ye, and Xiaojuan Qi.
\newblock Prior-free category-level pose estimation with implicit space transformation.
\newblock 2023{\natexlab{a}}.

\bibitem[Liu et~al.(2021)Liu, Jiang, Xu, Liu, and Wang]{liu2021semi}
Shaowei Liu, Hanwen Jiang, Jiarui Xu, Sifei Liu, and Xiaolong Wang.
\newblock Semi-supervised 3d hand-object poses estimation with interactions in time.
\newblock In \emph{CVPR}, 2021.

\bibitem[Liu et~al.(2023{\natexlab{b}})Liu, Hou, Yang, Li, and Lu]{liu2023revisit}
Xinpeng Liu, Haowen Hou, Yanchao Yang, Yong-Lu Li, and Cewu Lu.
\newblock Revisit human-scene interaction via space occupancy.
\newblock \emph{arXiv preprint arXiv:2312.02700}, 2023{\natexlab{b}}.

\bibitem[Liu et~al.(2022)Liu, Liu, Jiang, Lyu, Wan, Shen, Liang, Fu, Wang, and Yi]{liu2022hoi4d}
Yunze Liu, Yun Liu, Che Jiang, Kangbo Lyu, Weikang Wan, Hao Shen, Boqiang Liang, Zhoujie Fu, He Wang, and Li Yi.
\newblock Hoi4d: A 4d egocentric dataset for category-level human-object interaction.
\newblock \emph{CVPR}, 2022.

\bibitem[Liu et~al.(2024)Liu, Yang, Si, Liu, Li, Zhang, Liu, and Yi]{liu2024taco}
Yun Liu, Haolin Yang, Xu Si, Ling Liu, Zipeng Li, Yuxiang Zhang, Yebin Liu, and Li Yi.
\newblock Taco: Benchmarking generalizable bimanual tool-action-object understanding.
\newblock pages 21740--21751, 2024.

\bibitem[Luo et~al.(2024)Luo, Cao, Christen, Winkler, Kitani, and Xu]{luo2024grasping}
Zhengyi Luo, Jinkun Cao, Sammy Christen, Alexander Winkler, Kris Kitani, and Weipeng Xu.
\newblock Grasping diverse objects with simulated humanoids.
\newblock \emph{arXiv preprint arXiv:2407.11385}, 2024.

\bibitem[Mahmood et~al.(2019)Mahmood, Ghorbani, Troje, Pons-Moll, and Black]{mahmood2019amass}
Naureen Mahmood, Nima Ghorbani, Nikolaus~F Troje, Gerard Pons-Moll, and Michael~J Black.
\newblock Amass: Archive of motion capture as surface shapes.
\newblock \emph{ICCV}, 2019.

\bibitem[{Manus}(2023)]{manus}
{Manus}, 2023.
\newblock \url{https://www.manus-meta.com/}.

\bibitem[Matsuyama and Takai(2002)]{Matsuyama-2002}
T. Matsuyama and T. Takai.
\newblock Generation, visualization, and editing of 3d video.
\newblock \emph{3DPVT}, 2002.

\bibitem[Matusik et~al.(2000)Matusik, Buehler, Raskar, Gortler, and McMillan]{Matusik-2000}
Wojciech Matusik, Chris Buehler, Ramesh Raskar, Steven~J. Gortler, and Leonard McMillan.
\newblock Image-based visual hulls.
\newblock \emph{SIGGRAPH}, 2000.

\bibitem[Mir et~al.(2023)Mir, Puig, Kanazawa, and Pons-Moll]{mir2023generating}
Aymen Mir, Xavier Puig, Angjoo Kanazawa, and Gerard Pons-Moll.
\newblock Generating continual human motion in diverse 3d scenes.
\newblock \emph{arXiv preprint arXiv:2304.02061}, 2023.

\bibitem[{Movella}(2023)]{xsens}
{Movella}, 2023.
\newblock \url{https://base.xsens.com/}.

\bibitem[Nie et~al.(2022)Nie, Dai, Han, and Nießner]{nie2022pose2room}
Yinyu Nie, Angela Dai, Xiaoguang Han, and Matthias Nießner.
\newblock Pose2room: Understanding 3d scenes from human activities, 2022.

\bibitem[Pan et~al.(2023)Pan, Ma, Yi, Hu, Wang, Zhou, Li, and Xu]{pan2023fusing}
Shaohua Pan, Qi Ma, Xinyu Yi, Weifeng Hu, Xiong Wang, Xingkang Zhou, Jijunnan Li, and Feng Xu.
\newblock Fusing monocular images and sparse imu signals for real-time human motion capture.
\newblock In \emph{SIGGRAPH Asia}, pages 1--11, 2023.

\bibitem[Pavlakos et~al.(2019)Pavlakos, Choutas, Ghorbani, Bolkart, Osman, Tzionas, and Black]{pavlakos2019expressive}
Georgios Pavlakos, Vasileios Choutas, Nima Ghorbani, Timo Bolkart, Ahmed~AA Osman, Dimitrios Tzionas, and Michael~J Black.
\newblock Expressive body capture: 3d hands, face, and body from a single image.
\newblock In \emph{CVPR}, 2019.

\bibitem[Pavlakos et~al.(2024)Pavlakos, Shan, Radosavovic, Kanazawa, Fouhey, and Malik]{pavlakos2024reconstructing}
Georgios Pavlakos, Dandan Shan, Ilija Radosavovic, Angjoo Kanazawa, David Fouhey, and Jitendra Malik.
\newblock Reconstructing hands in 3d with transformers.
\newblock 2024.

\bibitem[Peng et~al.(2023)Peng, Xie, Wu, Jampani, Sun, and Jiang]{peng2023hoi}
Xiaogang Peng, Yiming Xie, Zizhao Wu, Varun Jampani, Deqing Sun, and Huaizu Jiang.
\newblock Hoi-diff: Text-driven synthesis of 3d human-object interactions using diffusion models.
\newblock \emph{arXiv preprint arXiv:2312.06553}, 2023.

\bibitem[Petit et~al.(2009)Petit, Lesage, Boyer, and Raffin]{Petit-2009}
Benjamin Petit, Jean-Denis Lesage, Edmond Boyer, and Bruno Raffin.
\newblock {Virtualization Gate}.
\newblock \emph{SIGGRAPH Emerging Technologies}, 2009.

\bibitem[Petrov et~al.(2023)Petrov, Marin, Chibane, and Pons-Moll]{petrov2023object}
Ilya~A Petrov, Riccardo Marin, Julian Chibane, and Gerard Pons-Moll.
\newblock Object pop-up: Can we infer 3d objects and their poses from human interactions alone?
\newblock \emph{CVPR}, 2023.

\bibitem[Petrovich et~al.(2022)Petrovich, Black, and Varol]{petrovich2022temos}
Mathis Petrovich, Michael~J Black, and G{\"u}l Varol.
\newblock Temos: Generating diverse human motions from textual descriptions.
\newblock \emph{ECCV}, 2022.

\bibitem[Plankers and Fua(2003)]{Plankers-03}
Ralf Plankers and Pascal Fua.
\newblock {Articulated Soft Objects for Multi-View Shape and Motion Capture}.
\newblock \emph{TPAMI}, 2003.

\bibitem[Pons-Moll et~al.(2023)Pons-Moll, Guzov, Chibane, Marin, He, and Sattler]{pons2023interaction}
Gerard Pons-Moll, Vladimir Guzov, Julian Chibane, Riccardo Marin, Yannan He, and Torsten Sattler.
\newblock Interaction replica: Tracking human-object interaction and scene changes from human motion.
\newblock 2023.

\bibitem[Schonberger and Frahm(2016)]{schonberger2016structure}
Johannes~L Schonberger and Jan-Michael Frahm.
\newblock Structure-from-motion revisited.
\newblock In \emph{CVPR}, 2016.

\bibitem[Sener et~al.(2022)Sener, Chatterjee, Shelepov, He, Singhania, Wang, and Yao]{sener2022assembly101}
Fadime Sener, Dibyadip Chatterjee, Daniel Shelepov, Kun He, Dipika Singhania, Robert Wang, and Angela Yao.
\newblock Assembly101: A large-scale multi-view video dataset for understanding procedural activities.
\newblock \emph{CVPR}, 2022.

\bibitem[Shimada et~al.(2022)Shimada, Golyanik, Li, P{\'e}rez, Xu, and Theobalt]{shimada2022hulc}
Soshi Shimada, Vladislav Golyanik, Zhi Li, Patrick P{\'e}rez, Weipeng Xu, and Christian Theobalt.
\newblock Hulc: 3d human motion capture with pose manifold sampling and dense contact guidance.
\newblock In \emph{ECCV}. Springer, 2022.

\bibitem[Sigal et~al.(2010)Sigal, Balan, and Black]{sigal2010humaneva}
Leonid Sigal, Alexandru~O Balan, and Michael~J Black.
\newblock Humaneva: Synchronized video and motion capture dataset and baseline algorithm for evaluation of articulated human motion.
\newblock \emph{IJCV}, 87\penalty0 (1-2):\penalty0 4, 2010.

\bibitem[Song et~al.(2020)Song, Meng, and Ermon]{song2020denoising}
Jiaming Song, Chenlin Meng, and Stefano Ermon.
\newblock Denoising diffusion implicit models.
\newblock \emph{arXiv:2010.02502}, 2020.

\bibitem[Stoll et~al.(2011)Stoll, Hasler, Gall, Seidel, and Theobalt]{Stoll-11}
Carsten Stoll, Nils Hasler, Juergen Gall, Hans-Peter Seidel, and Christian Theobalt.
\newblock Fast articulated motion tracking using a sums of gaussians body model.
\newblock \emph{ICCV}, 2011.

\bibitem[Taheri et~al.(2020)Taheri, Ghorbani, Black, and Tzionas]{taheri2020grab}
Omid Taheri, Nima Ghorbani, Michael~J Black, and Dimitrios Tzionas.
\newblock Grab: A dataset of whole-body human grasping of objects.
\newblock \emph{ECCV}, 2020.

\bibitem[Taheri et~al.(2022)Taheri, Choutas, Black, and Tzionas]{taheri2022goal}
Omid Taheri, Vasileios Choutas, Michael~J Black, and Dimitrios Tzionas.
\newblock Goal: Generating 4d whole-body motion for hand-object grasping.
\newblock In \emph{CVPR}, 2022.

\bibitem[Tendulkar et~al.(2023)Tendulkar, Sur{\'\i}s, and Vondrick]{tendulkar2023flex}
Purva Tendulkar, D{\'\i}dac Sur{\'\i}s, and Carl Vondrick.
\newblock Flex: Full-body grasping without full-body grasps.
\newblock \emph{CVPR}, 2023.

\bibitem[Tevet et~al.(2022)Tevet, Raab, Gordon, Shafir, Cohen-Or, and Bermano]{tevet2022human}
Guy Tevet, Sigal Raab, Brian Gordon, Yonatan Shafir, Daniel Cohen-Or, and Amit~H Bermano.
\newblock Human motion diffusion model.
\newblock \emph{arXiv preprint arXiv:2209.14916}, 2022.

\bibitem[Turpin et~al.(2022)Turpin, Wang, Heiden, Chen, Macklin, Tsogkas, Dickinson, and Garg]{turpin2022grasp}
Dylan Turpin, Liquan Wang, Eric Heiden, Yun-Chun Chen, Miles Macklin, Stavros Tsogkas, Sven Dickinson, and Animesh Garg.
\newblock Grasp’d: Differentiable contact-rich grasp synthesis for multi-fingered hands.
\newblock \emph{ECCV}, 2022.

\bibitem[Vlasic et~al.(2008{\natexlab{a}})Vlasic, Baran, Matusik, and Popovi{\'{c}}]{Vlasic-08}
Daniel Vlasic, Ilya Baran, Wojciech Matusik, and Jovan Popovi{\'{c}}.
\newblock {Articulated mesh animation from multi-view silhouettes}.
\newblock \emph{TOG}, 2008{\natexlab{a}}.

\bibitem[Vlasic et~al.(2008{\natexlab{b}})Vlasic, Baran, Matusik, and Popovi\'{c}]{Vlasic-2008}
Daniel Vlasic, Ilya Baran, Wojciech Matusik, and Jovan Popovi\'{c}.
\newblock Articulated mesh animation from multi-view silhouettes.
\newblock \emph{SIGGRAPH}, 2008{\natexlab{b}}.

\bibitem[Wang et~al.(2021{\natexlab{a}})Wang, Xu, Xu, Liu, and Wang]{wang2021synthesizing}
Jiashun Wang, Huazhe Xu, Jingwei Xu, Sifei Liu, and Xiaolong Wang.
\newblock Synthesizing long-term 3d human motion and interaction in 3d scenes.
\newblock In \emph{CVPR}, 2021{\natexlab{a}}.

\bibitem[Wang et~al.(2021{\natexlab{b}})Wang, Yan, Dai, and Lin]{wang2020motion}
Jingbo Wang, Sijie Yan, Bo Dai, and Dahua Lin.
\newblock Scene-aware generative network for human motion synthesis.
\newblock In \emph{CVPR}, 2021{\natexlab{b}}.

\bibitem[Wang et~al.(2022{\natexlab{a}})Wang, Rong, Liu, Yan, Lin, and Dai]{wang2022towards}
Jingbo Wang, Yu Rong, Jingyuan Liu, Sijie Yan, Dahua Lin, and Bo Dai.
\newblock Towards diverse and natural scene-aware 3d human motion synthesis.
\newblock In \emph{CVPR}, 2022{\natexlab{a}}.

\bibitem[Wang et~al.(2022{\natexlab{b}})Wang, Li, Kuo, Kocabas, Aksan, and Hilliges]{wang2022reconstructing}
Xi Wang, Gen Li, Yen-Ling Kuo, Muhammed Kocabas, Emre Aksan, and Otmar Hilliges.
\newblock Reconstructing action-conditioned human-object interactions using commonsense knowledge priors.
\newblock In \emph{3DV}. IEEE, 2022{\natexlab{b}}.

\bibitem[Wang et~al.(2023)Wang, Lin, Zeng, Luo, Zhang, and Zhang]{wang2023physhoi}
Yinhuai Wang, Jing Lin, Ailing Zeng, Zhengyi Luo, Jian Zhang, and Lei Zhang.
\newblock Physhoi: Physics-based imitation of dynamic human-object interaction.
\newblock \emph{arXiv preprint arXiv:2312.04393}, 2023.

\bibitem[Wang et~al.(2022{\natexlab{c}})Wang, Chen, Liu, Zhu, Liang, and Huang]{wang2022humanise}
Zan Wang, Yixin Chen, Tengyu Liu, Yixin Zhu, Wei Liang, and Siyuan Huang.
\newblock Humanise: Language-conditioned human motion generation in 3d scenes.
\newblock \emph{NeurIPS}, 2022{\natexlab{c}}.

\bibitem[Xie et~al.(2023)Xie, Bhatnagar, and Pons-Moll]{xie2023visibility}
Xianghui Xie, Bharat~Lal Bhatnagar, and Gerard Pons-Moll.
\newblock Visibility aware human-object interaction tracking from single rgb camera.
\newblock In \emph{CVPR}, 2023.

\bibitem[Xie et~al.(2024)Xie, Lenssen, and Pons-Moll]{xie2024intertrack}
Xianghui Xie, Jan~Eric Lenssen, and Gerard Pons-Moll.
\newblock Intertrack: Tracking human object interaction without object templates.
\newblock \emph{arXiv preprint arXiv:2408.13953}, 2024.

\bibitem[Xu et~al.(2023)Xu, Li, Wang, and Gui]{xu2023interdiff}
Sirui Xu, Zhengyuan Li, Yu-Xiong Wang, and Liang-Yan Gui.
\newblock Interdiff: Generating 3d human-object interactions with physics-informed diffusion.
\newblock \emph{ICCV}, 2023.

\bibitem[Xu et~al.(2024)Xu, Wang, Wang, and Gui]{xu2024interdreamer}
Sirui Xu, Ziyin Wang, Yu-Xiong Wang, and Liang-Yan Gui.
\newblock Interdreamer: Zero-shot text to 3d dynamic human-object interaction.
\newblock \emph{arXiv preprint arXiv:2403.19652}, 2024.

\bibitem[Yang et~al.(2024)Yang, Niu, Jiang, Zhang, and Huang]{yang2024f}
Jie Yang, Xuesong Niu, Nan Jiang, Ruimao Zhang, and Siyuan Huang.
\newblock F-hoi: Toward fine-grained semantic-aligned 3d human-object interactions.
\newblock \emph{arXiv preprint arXiv:2407.12435}, 2024.

\bibitem[Yang et~al.(2021)Yang, Zhan, Li, Xu, Li, and Lu]{yang2021cpf}
Lixin Yang, Xinyu Zhan, Kailin Li, Wenqiang Xu, Jiefeng Li, and Cewu Lu.
\newblock Cpf: Learning a contact potential field to model the hand-object interaction.
\newblock In \emph{ICCV}, 2021.

\bibitem[Yi et~al.(2023)Yi, Huang, Tripathi, Hering, Thies, and Black]{yi2023mime}
Hongwei Yi, Chun-Hao~P Huang, Shashank Tripathi, Lea Hering, Justus Thies, and Michael~J Black.
\newblock Mime: Human-aware 3d scene generation.
\newblock In \emph{CVPR}, pages 12965--12976, 2023.

\bibitem[Yi et~al.(2024)Yi, Thies, Black, Peng, and Rempe]{yi2024tesmo}
Hongwei Yi, Justus Thies, Michael~J. Black, Xue~Bin Peng, and Davis Rempe.
\newblock Generating human interaction motions in scenes with text control.
\newblock \emph{ECCV}, 2024.

\bibitem[Zakour et~al.(2024)Zakour, Nath, Lohmer, G{\"o}k{\c{c}}e, Piccolrovazzi, Patsch, Wu, Chaudhari, and Steinbach]{zakour2024adl4d}
Marsil Zakour, Partha~Pratim Nath, Ludwig Lohmer, Emre~Faik G{\"o}k{\c{c}}e, Martin Piccolrovazzi, Constantin Patsch, Yuankai Wu, Rahul Chaudhari, and Eckehard Steinbach.
\newblock Adl4d: Towards a contextually rich dataset for 4d activities of daily living.
\newblock \emph{arXiv preprint arXiv:2402.17758}, 2024.

\bibitem[Zhan et~al.(2024)Zhan, Yang, Zhao, Mao, Xu, Lin, Li, and Lu]{zhan2024oakink2}
Xinyu Zhan, Lixin Yang, Yifei Zhao, Kangrui Mao, Hanlin Xu, Zenan Lin, Kailin Li, and Cewu Lu.
\newblock Oakink2: A dataset of bimanual hands-object manipulation in complex task completion.
\newblock 2024.

\bibitem[Zhang et~al.(2021)Zhang, Ye, Shiratori, and Komura]{zhang2021manipnet}
He Zhang, Yuting Ye, Takaaki Shiratori, and Taku Komura.
\newblock Manipnet: Neural manipulation synthesis with a hand-object spatial representation.
\newblock \emph{ACM TOG}, 40\penalty0 (4), 2021.

\bibitem[Zhang et~al.(2023{\natexlab{a}})Zhang, Christen, Fan, Zheng, Hwangbo, Song, and Hilliges]{zhang2023artigrasp}
Hui Zhang, Sammy Christen, Zicong Fan, Luocheng Zheng, Jemin Hwangbo, Jie Song, and Otmar Hilliges.
\newblock Artigrasp: Physically plausible synthesis of bi-manual dexterous grasping and articulation.
\newblock \emph{arXiv preprint arXiv:2309.03891}, 2023{\natexlab{a}}.

\bibitem[Zhang et~al.(2023{\natexlab{b}})Zhang, Luo, Yang, Xu, Wu, Shi, Yu, Xu, and Wang]{zhang2023neuraldome}
Juze Zhang, Haimin Luo, Hongdi Yang, Xinru Xu, Qianyang Wu, Ye Shi, Jingyi Yu, Lan Xu, and Jingya Wang.
\newblock Neuraldome: A neural modeling pipeline on multi-view human-object interactions.
\newblock In \emph{CVPR}, 2023{\natexlab{b}}.

\bibitem[Zhang et~al.(2023{\natexlab{c}})Zhang, Zhang, Zhang, Zhou, Zhou, Shao, Hu, and Liu]{zhang2023ins}
Jiajun Zhang, Yuxiang Zhang, Hongwen Zhang, Xiao Zhou, Boyao Zhou, Ruizhi Shao, Zonghai Hu, and Yebin Liu.
\newblock Ins-hoi: Instance aware human-object interactions recovery.
\newblock \emph{arXiv preprint arXiv:2312.09641}, 2023{\natexlab{c}}.

\bibitem[Zhang et~al.(2024{\natexlab{a}})Zhang, Zhang, Song, Shi, Zhao, Shi, Yu, Xu, and Wang]{zhang2024hoi}
Juze Zhang, Jingyan Zhang, Zining Song, Zhanhe Shi, Chengfeng Zhao, Ye Shi, Jingyi Yu, Lan Xu, and Jingya Wang.
\newblock Hoi-m\^{} 3: Capture multiple humans and objects interaction within contextual environment.
\newblock 2024{\natexlab{a}}.

\bibitem[Zhang et~al.(2024{\natexlab{b}})Zhang, Zhang, An, Li, Zhang, Hu, and Liu]{zhang2024manidext}
Jiajun Zhang, Yuxiang Zhang, Liang An, Mengcheng Li, Hongwen Zhang, Zonghai Hu, and Yebin Liu.
\newblock Manidext: Hand-object manipulation synthesis via continuous correspondence embeddings and residual-guided diffusion.
\newblock \emph{arXiv preprint arXiv:2409.09300}, 2024{\natexlab{b}}.

\bibitem[Zhang et~al.(2022{\natexlab{a}})Zhang, Cai, Pan, Hong, Guo, Yang, and Liu]{zhang2022motiondiffuse}
Mingyuan Zhang, Zhongang Cai, Liang Pan, Fangzhou Hong, Xinying Guo, Lei Yang, and Ziwei Liu.
\newblock Motiondiffuse: Text-driven human motion generation with diffusion model.
\newblock \emph{arXiv preprint arXiv:2208.15001}, 2022{\natexlab{a}}.

\bibitem[Zhang et~al.(2025)Zhang, Wu, Wang, Wang, Liu, and Lin]{zhang2025omni6d}
Mengchen Zhang, Tong Wu, Tai Wang, Tengfei Wang, Ziwei Liu, and Dahua Lin.
\newblock Omni6d: Large-vocabulary 3d object dataset for category-level 6d object pose estimation.
\newblock 2025.

\bibitem[Zhang et~al.(2022{\natexlab{b}})Zhang, Bhatnagar, Starke, Guzov, and Pons-Moll]{zhang2022couch}
Xiaohan Zhang, Bharat~Lal Bhatnagar, Sebastian Starke, Vladimir Guzov, and Gerard Pons-Moll.
\newblock Couch: Towards controllable human-chair interactions.
\newblock In \emph{ECCV}, 2022{\natexlab{b}}.

\bibitem[Zheng et~al.(2023{\natexlab{a}})Zheng, Zheng, Fang, Liu, and Yi]{zheng2023cams}
Juntian Zheng, Qingyuan Zheng, Lixing Fang, Yun Liu, and Li Yi.
\newblock Cams: Canonicalized manipulation spaces for category-level functional hand-object manipulation synthesis.
\newblock \emph{CVPR}, 2023{\natexlab{a}}.

\bibitem[Zheng et~al.(2023{\natexlab{b}})Zheng, Shi, Cui, Zhao, Luo, and Zhou]{zheng2023coop}
Yanzhao Zheng, Yunzhou Shi, Yuhao Cui, Zhongzhou Zhao, Zhiling Luo, and Wei Zhou.
\newblock Coop: Decoupling and coupling of whole-body grasping pose generation.
\newblock pages 2163--2173, 2023{\natexlab{b}}.

\bibitem[Zhou et~al.(2022)Zhou, Bhatnagar, Lenssen, and Pons-Moll]{zhou2022toch}
Keyang Zhou, Bharat~Lal Bhatnagar, Jan~Eric Lenssen, and Gerard Pons-Moll.
\newblock Toch: Spatio-temporal object-to-hand correspondence for motion refinement.
\newblock \emph{ECCV}, 2022.

\bibitem[Zhou et~al.(2024)Zhou, Bhatnagar, Lenssen, and Pons-Moll]{zhou2024gears}
Keyang Zhou, Bharat~Lal Bhatnagar, Jan~Eric Lenssen, and Gerard Pons-Moll.
\newblock Gears: Local geometry-aware hand-object interaction synthesis.
\newblock pages 20634--20643, 2024.

\bibitem[Zhou et~al.(2019)Zhou, Barnes, Jingwan, Jimei, and Hao]{zhou20196d}
Yi Zhou, Connelly Barnes, Lu Jingwan, Yang Jimei, and Li Hao.
\newblock On the continuity of rotation representations in neural networks.
\newblock In \emph{CVPR}, 2019.

\bibitem[Zimmermann et~al.(2019)Zimmermann, Ceylan, Yang, Russell, Argus, and Brox]{zimmermann2019freihand}
Christian Zimmermann, Duygu Ceylan, Jimei Yang, Bryan Russell, Max Argus, and Thomas Brox.
\newblock Freihand: A dataset for markerless capture of hand pose and shape from single rgb images.
\newblock \emph{ICCV}, 2019.

\bibitem[Zoss et~al.(2018)Zoss, Bradley, B{\'e}rard, and Beeler]{zoss2018empirical}
Gaspard Zoss, Derek Bradley, Pascal B{\'e}rard, and Thabo Beeler.
\newblock An empirical rig for jaw animation.
\newblock \emph{ACM TOG}, 37\penalty0 (4):\penalty0 1--12, 2018.

\end{thebibliography}
}

\clearpage 
\appendix
\label{sec:appendix}

\begin{figure}
    \centering
    \includegraphics[width=\linewidth, trim={0cm 0.0cm 0cm 0.2cm},clip]{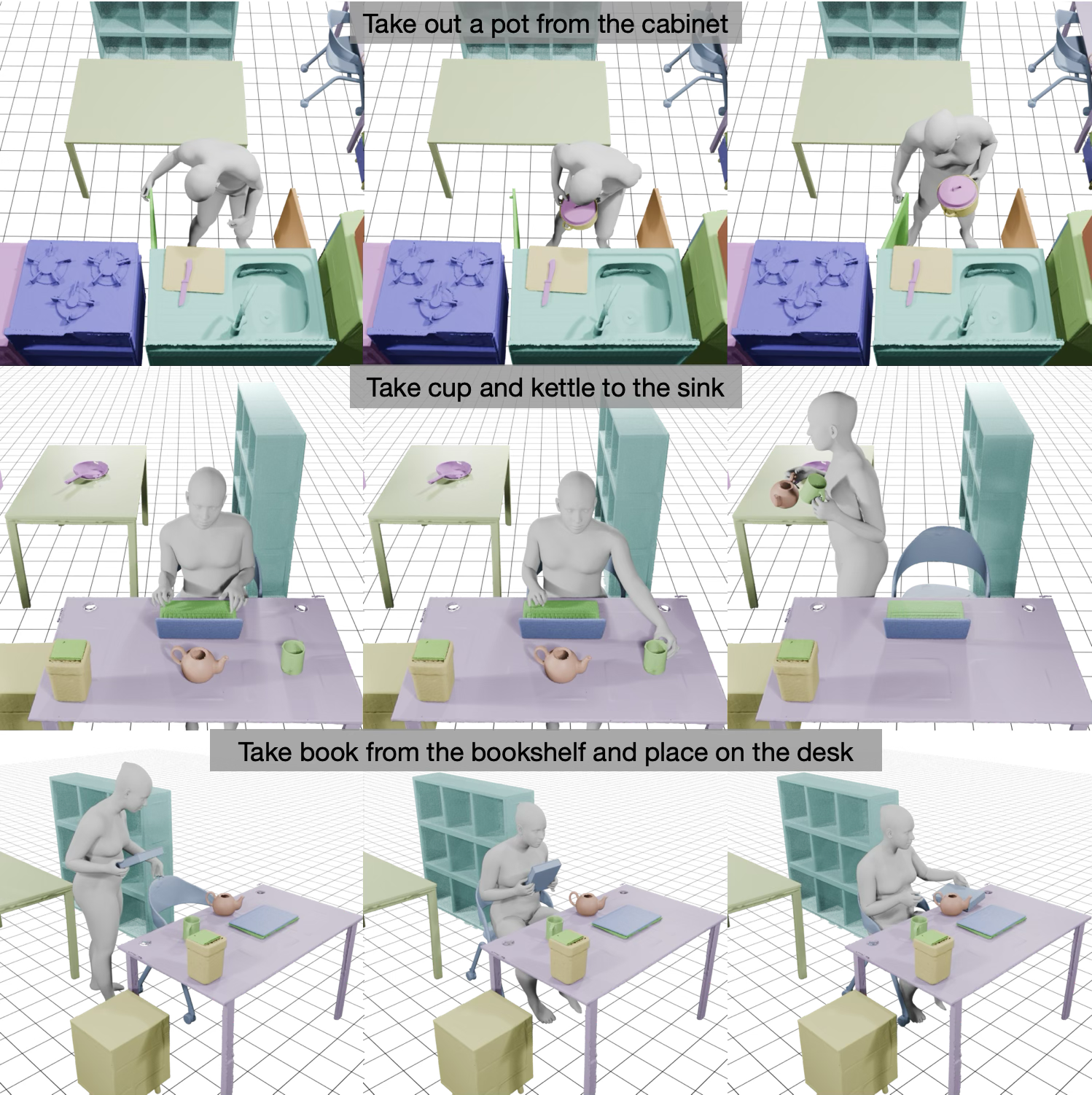}
    \caption{Rendered scenes and text annotation for each scene from example scenario.}
    \label{fig:example_scenario}
    \vspace{-10pt}
\end{figure}

\section{Dataset Details} 
\subsection{Dataset Contents}
\noindent \textbf{Scanned Object Mesh.}  We obtain high-quality 3D mesh scans of all objects placed in our system via an Einstar3D scanner~\cite{einstar}. We scan each object at least twice to reduce the unscanned areas or holes by changing the orientations of the objects (e.g., up-side-down), and fuse the scanned meshes via manual alignments. They are zero-centered and scaled to a metric scale.

\noindent \textbf{Object Articulation Information.}  
Objects with articulation contain axis $a_e$. If the part has a revolute joint, we include pivot point $p_e$ additionally. These are defined in the object canonical space and are utilized in getting each object part-transformation toward the camera space.

\noindent \textbf{Object Position and Orientation in the Camera Space.} Each object's spatial information is computed using the per-frame transformation of markers attached to each object.

\noindent \textbf{Relative Orientation of Hand/Body Joints.}  
Orientation of each hand and body joints with respect to their parent joints is recorded and processed via a motion capture system.

\noindent \textbf{3D Hand/Body Joint Positions in the Camera Space.}  
With the positions of markers attached to the body in the mocap space acquired via body alignment protocol, translation and orientation of body to camera space are obtained using the positions of corresponding markers in the camera spaces. We compute the positions of two hands and body using the obtained translation and orientation.

\noindent \textbf{Text Annotation for Each Action.}  
For each capture, participants receive verbal instructions detailing the actions they will perform. These instructions specify which objects to interact with and how to interact with them, as illustrated in Fig.~\ref{fig:example_scenario}. The instructions are recorded and synchronized with the motion data. Additionally, we manually inspect the instrument to create more accurate text annotations, ensuring they are reliably mapped to each action.

\noindent \textbf{Per-frame Contact Information.}  
At each frame where contact between Left/Right/Body and object occurs, the corresponding frame and object category/body part information is recorded.

\subsection{Dataset Comparison}

\begin{table*}%
  \centering
  \resizebox{\textwidth}{!}{
  \begin{tabular}{c|ccccccccccc}
    \toprule
    Dataset & hours \#  & subject \# & object \# & body & hand \#& contact& {obj. 6d.} &{obj. artic.} &  {multi obj.}  & setup \\
    \midrule
     {GRAB { \cite{taheri2020grab}}}& 3.8 & 10 &51 & \cmark  & 2 & \cmark & \cmark & \xmark & \xmark &  { standing}\\
     {BEHAVE { \cite{bhatnagar2022behave}}} & 4.2 & 8 & 20  & \cmark  & - & \cmark& \cmark & \xmark & \xmark &  {portable}\\
     {InterCap { \cite{huang2022intercap}}} & 0.6 & 10 &10& \cmark  & 2 & \cmark& \cmark & \xmark & \xmark &  {portable}\\
     {FHPA { \cite{garcia2018first}}}& 0.9 & 6 &26 & \xmark  & 1 & \xmark& \cmark & \xmark & \xmark  &  {room}\\
     {H2O { \cite{kwon2021h2o}}} & 1.1 & 4 & 8 & \xmark & 2 & \xmark& \cmark & \xmark & \xmark  &  {table}\\
     {H2O-3D { \cite{hampali2022keypoint}}} & - & 5 &10 & \xmark  & 2& \xmark & \cmark & \xmark & \xmark  &  {table}\\
     {HOI4D { \cite{liu2022hoi4d}}}& 22.2 & 9 & 800(16) & \xmark & 1 & \xmark& \cmark & \cmark & \xmark  &  {room}\\
     {Chairs { \cite{jiang2023full}}}& 17.3 & 46 & 81 & \cmark & 2 & - & \cmark & \cmark & \xmark  &  {standing}\\
     {ARCTIC { \cite{fan2023arctic}}}& 1.2 & 10 &11 & \cmark  & 2 & \cmark& \cmark & \cmark & \xmark &  {standing}\\
     {NeuralDome { \cite{zhang2023neuraldome}}}& 4.6 & 10 & 23 & \cmark & 2 & \cmark& \cmark & \cmark & \xmark  &  {standing}\\
     {OAKINK2 { \cite{zhan2024oakink2}}}& 12.38 & 9 & 75 & \cmark  & 2 & \cmark& \cmark & \cmark & \cmark &  {table}\\
     {TACO { \cite{liu2024taco}}}& 2.53 & 14 &196(20) & \xmark  & 2 & \cmark& \cmark & \xmark & \cmark &  {table}\\
     {TRUMANS { \cite{jiang2024scaling}}}& 15 & 7 & 20(placeholders) & \cmark  & 2 & \cmark& \cmark & \cmark & \cmark &  {room}\\
    \midrule
    \textbf{Ours} & 8.1 & 38 &22& \cmark & 2 & \cmark & \cmark & \cmark &  \cmark &   {room} \\
    \bottomrule
  \end{tabular}%
  }
    \caption{Comparison of existing human-object interaction datasets}
    \label{tab:comparison-db}
\end{table*} %

As shown, our ParaHome dataset is the comprehensive dataset which capture all authentic and dynamic human-object interaction scenarios in a natural room environment. Our dataset includes  dexterous body motion and movement of all objects in the scene and encompasses natural manipulation motions involving articulated objects and multiple objects even in concurrent usage scenarios. Our capture scenarios feature natural and sequential manipulations like cooking, as shown in our supplementary video. Furthermore, we collected data from 38 participants, capturing a wide range of motion styles across individuals.

\section{System Details}

\begin{figure}[t]
    \centering
    \includegraphics[width=\linewidth, trim={0cm 0.0cm 0cm 0cm},clip]{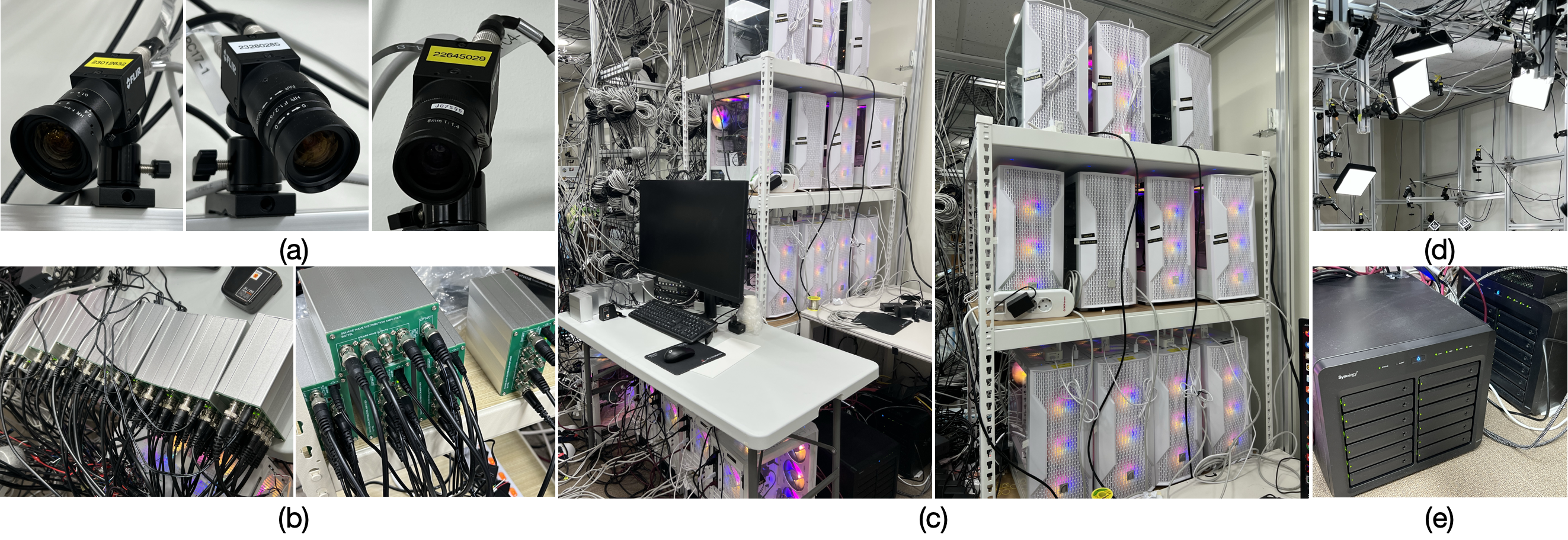}
    \caption{System Devices. (a) RGB cameras with 3 types of lenses. (b) Signal distributers (c) Desktop machines (d) LED Lights (e) NAS storage systems}
    \label{fig:supp_system}
    \vspace{-10pt}
\end{figure}

\subsection{Using ArUco Markers}
Even though several works~\cite{jiang2023full, fan2023arctic} utilized IR markers for motion tracking, we find using ArUco markers to be more suitable in our capture system. We aim to capture in a broader spatial spectrum(i.e. entire room setting filled with objects) involving multiple interactions in a single capture time.
Such environments filled with multiple furniture/objects and hand-object interactions involving multiple direct contacts, cause a significant occlusion as simulated in Sec.4.1 of our main manuscript. %
Even though ArUco markers have its downsides in corrupting RGB data and influencing natural human motions, using RGB data is not within our interest as mentioned Sec 6. of main manuscript and we empirically placed markers to minimize such interruptions.

\subsection{Hardware Details}
In order to cover the entire volume of the room and to reduce occlusion issues, we install 70 RGB industrial cameras, BFLY-31S4C-C. 
The cameras capture videos at 30Hz in 2048$\times$1536 resolution.
We set the exposure time at 3$msec$, which shows a good balance between low-motion blur and sufficient brightness. 
We use three types of lenses (thirty 3$mm$ lenses, twenty 5$mm$ lenses, and twenty 6$mm$ lenses), where the wide-angle lens (3$mm$) is helpful in capturing wide area. 
We calibrate the cameras using Structure-from-Motion via COLMAP~\cite{schonberger2016structure} with multiple randomly patterned fabrics placed in our system. We provide pre-calibrated initial intrinsic parameters for the three types of lenses derived from 2 or 3 samples of lenses for better convergence in camera pose estimation. 
We scale the calibrated 3D space into a real-world metric (in meters) by locating checkerboards with known sizes during camera calibration.

All cameras, the motion capture suit, and gloves are synced and gen-locked via a common square wave signal that comes from the motion capture device to synchronize two heterogeneous systems, which is crucial to precise HOI captures. %
To deliver the sync signals to a large number of cameras, we utilized 11 signal distributors in a hierarchical manner, each of which can be connected to 8 cameras via GPIO cables. We use 1 master and 18 slave desktop machines to control the cameras and process captured records. Each slave machine is connected to 3 or 4 cameras via Ethernet cables and equipped with a 4-port 1G ethernet board, and 2 SSDs with a capacity of 500GB and 1TB each. 15 LED lights (4500$lm$) are installed to provide sufficient illumination. Pictures of our system devices are shown in Fig.~\ref{fig:supp_system}.

To capture both body motion and subtle hand motions, we use IMU-based motion capture equipments, Xsens motion suit~\cite{xsens} and Manus hand gloves~\cite{manus}. The body motion system captures the motions at 60Hz.

\begin{table}[h]
    \centering
    \begin{tabular}{c|cc}
    \toprule
 \multicolumn{1}{c}{Object} &   Part1 &  Part2 \\
    \midrule
        Sink   & revolute   & revolute        \\
        Laptop   & revolute     & -      \\
        Drawer   & sliding     & sliding      \\
        Gas stove  & revolute   & revolute     \\
        Microwave  & revolute    & -    \\
        Trashbin   & revolute   & -    \\
        Washing machine & revolute & - \\
        Refrigerator  & revolute  & revolute     \\
    \bottomrule
    \end{tabular}
    \caption{Part information of articulated objects }
    \label{tab:artinfo}
\end{table}

\section{Data Acquisition}
\subsection{Modeling Object Articulations.}

To capture the movement of articulated objects, we model each object as a parametric 3D model by defining the object-specific articulated motion parameters. This modeling requires scanning individual parts separately and compositing them in a canonical space by defining axis direction, pivot points, revolute joints, and so on, based on the object types. During HOI captures, we track the motion of each part via our marker system (e.g. monitor of a laptop and the base), from which we compute the articulated motion parameters. In this subsection, we describe the process of modeling articulated objects as parametric 3D models. Articulation information of each object with multiple parts is shown in Tab.~\ref{tab:artinfo}.

To find axis $\vb{a}_e$ and pivot point $\vb{p}_e$ of the articulated objects, we capture markers attached to each object part at different part states separately and acquire each marker corners in the \ParaHouse space as $\{m_i(t)\}_{t=1}^n$. Prior to applying algorithm, we transform marker corners $\{m_i(t)\}_{t=1}^n$ back to object canonical space with $T_{mar \rightarrow obj}^{-1}$ and utilize transformed marker corners in the canonical space $\{m_i^{\prime}(t)\}_{t=1}^n$. For the sliding joint, axis $a_e$ can easily be calculated using marker corners at time $t$ and $t^{\prime}$ as: 
 $$ \vb{a}_e = \frac{m_i^{\prime}(t) - m_i^{\prime}(t^{\prime})}{\|m_i^{\prime}(t) - m_i^{\prime}(t^{\prime})\|}$$

In case object part has a revolute joint, we start initializing an axis $\vb{a}_e$ and each relative state $ \Delta s_e(t, t^{\prime}) = |s_e(t) -s_e(t^{\prime})|$ between time $t$ and $t^{\prime}$ (for the target articulated object captured at different $n$ number of states, time $t$ and time $t^{\prime}$ satisfies $ t \neq t^{\prime}$ and $ t, t^{\prime} \in \{1,2,\cdots ,n \}$). Then we apply optimization with marker corners toward all possible pairs of times $t$ and $t^{\prime}$. Let $f$ be a map defining rotation transformation with respect to pivot and given axis-angle and denote as $\mb{T}_{t^{\prime} \rightarrow t} = f(a_e, \Delta s_e(t, t^{\prime}), p_e)$. Then for a set of all possible time pairs $\mathbf{P}$, the optimization target for axis $a_e$, relative state $\Delta s_e$ and pivot $p_e$ is defined as:
$$
\argmin_{a_e, \Delta s_e(t,t^{\prime}), p_e}\sum_{(t,t^{\prime}) \in \mathbf{P}}\|m_i^{\prime}(t) - \mb{T}_{t^{\prime} \rightarrow t}m_i^{\prime}(t^{\prime}) \|^2
$$
Since initial axis $\vb{a}_e$ and pivot $\vb{p}_e$ are initialized in the object canonical space, we directly utilize acquired information to derive transformations using detected markers for each capture data.

\subsection{Body Alignment Detail (Sec 3.4 in Main Paper)}

In this subsection, we provide additional details of our spatial alignment process between a multiview camera system and wearable motion capture systems, described in Sec. 3.2 in our main manuscript. 

To resolve the issue of imperfect body and hand skeleton scale from the wearable motion capture system, we attach 3 or 4 ArUco markers to each near-rigid body part (torso, hands, upper arms, lower arms, upper legs, lower legs) to assign correspondences. During alignment capture, participants perform the range-of-motion movement by rotating their arms and legs while pinned or bent, particularly twisting their wrists to locate each hand wrist. With the captured data, %
we optimize body skeleton configuration $\mathcal{B}=\{\mathcal{O} \}$ and body markers locations $\mathcal{M}^b$ via gradient decent with a learning rate of 0.008 for 50 epochs. Specifically for weights of body and foot, $\lambda_{b}=100,  \lambda_{f} = 5000$ are used. %
In the case that alignment is not well optimized, we additionally penalize excessive length change in spines and difference in skeleton lengths between the left and right sides of the body by adding an extra regularization term.
Once the alignment procedure is finished, 
we remove markers (all from the upper legs, one for each upper arm, lower arm, and lower leg) to minimize interference with the movements of the participant.  
The selection of remaining markers is determined based on their importance during captures, where we assess their importance by evaluating whether their absence would compromise the accuracy of body positioning in the camera space.
Check our supplementary video for an example of body alignment motion.

\subsection{Hand Calibration Structure and Protocol (Sec 3.4 in Main Paper)}

\begin{figure}[tb]
    \centering
    \begin{subfigure}{0.48\linewidth}
        \centering  %
        \includegraphics[width=\linewidth, trim={13cm 1cm 13cm 1cm},clip]{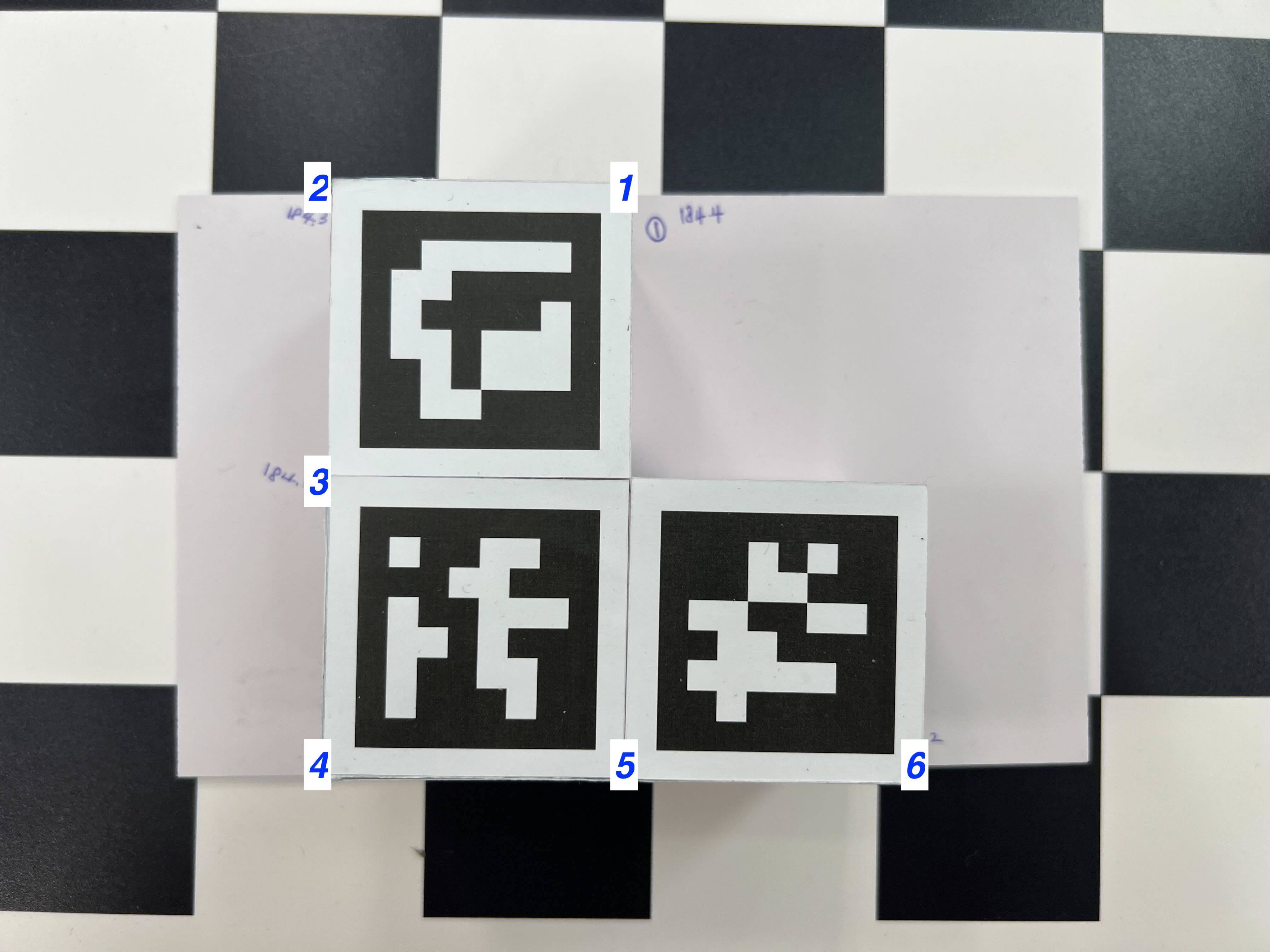}
    \end{subfigure}
    \hfill
    \begin{subfigure}{0.48\linewidth}
        \centering
        \includegraphics[width=\linewidth, trim={0cm 0.0cm 0cm 0cm},clip]{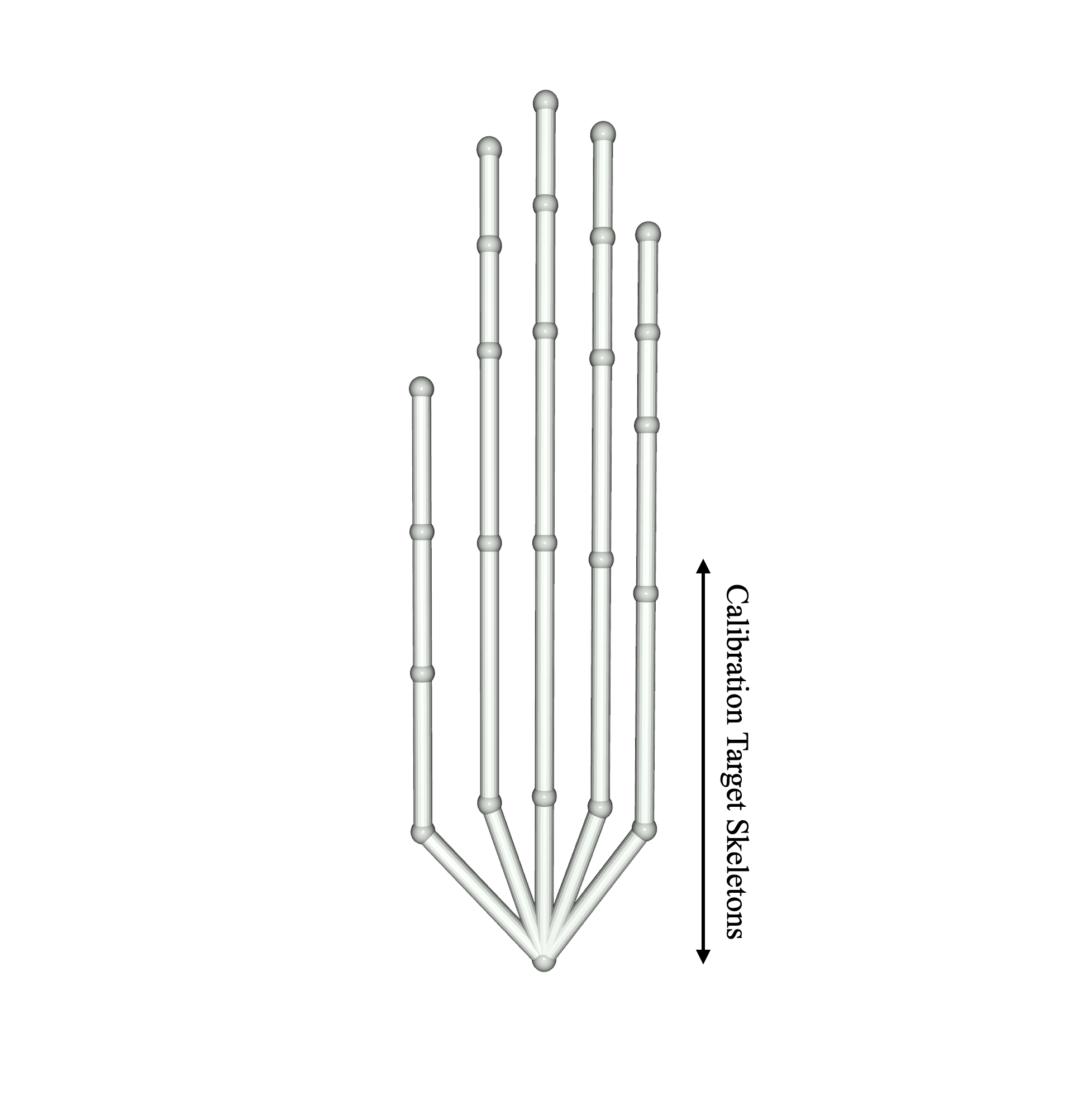}
    \end{subfigure}
    \caption{(Left) The hand calibration structure to precisely measure hand skeleton configuration and to find the relative locations of hand markers attached to the wrist in hand-centric coordinate (Right) Hand skeleton and Calibration targets}
    \label{fig:supp_combined}
\end{figure}

\begin{table}[tb]
\centering
    \resizebox{\columnwidth}{!}{
    \begin{tabular}{cccccc} 
    \toprule
    Corner \# &     Hand Side &     \makebox[2cm]{Seq1}      & \makebox[2cm]{Seq2}         & \makebox[2cm]{Seq3}           & \makebox[2cm]{Seq4} \\
    \midrule
        1,2   & Right     & 1, 2         & 1, 3       & 1, 4         & 1, 5 \\
        1,3   & Right     & 1, 2         & 1, 3       & 1, 4         & 1, 5 \\
        2,4   & Right     & 1, 2         & 1, 3       & 1, 4         & 1, 5 \\
        5,2   & Right     & 1, 2         & 1, 3       & 1, 4         & 1, 5 \\
        6,2   & Right     & 1, 2         & 1, 3       & 1, 4         & 1, 5 \\
        6,3,2 & Right     & 1, 2, 3 & 1, 3, 4 & 1, 4, 5 &      -   \\
    \midrule
        2,1   & Left      & 1, 2         & 1, 3       & 1, 4         & 1, 5 \\
        3,1   & Left      & 1, 2         & 1, 3       & 1, 4         & 1, 5 \\
        4,2   & Left      & 1, 2         & 1, 3       & 1, 4         & 1, 5 \\
        5,2   & Left      & 1, 2         & 1, 3       & 1, 4         & 1, 5 \\
        6,2   & Left      & 1, 2         & 1, 3       & 1, 4         & 1, 5 \\
        2,5,6 & Left      & 1, 2, 3  & 1, 3, 4  & 1, 4, 5  &      -   \\
    \bottomrule
    \end{tabular}
    }
    \caption{Hand Calibration Protocol}
    \label{tab:handcalib}
\end{table}

As human usually handle objects with their fingers, fingertips play an important role during interaction. We made the calibration structure to better locate fingertips and find the hand skeletons and relative locations between the attached hand markers to each wrist. The hand calibration structure is composed of three cubes with ArUco markers and the ordered 3D corner vertices of the structure are defined as $ C = \{\vb{c}_i \in \mathbb{R}^3 \}_{i=1}^{6}$ as shown in Fig.~\ref{fig:supp_combined}. During the hand calibration procedure, we request each participant to touch the calibration structure's corners with their fingertips. We instruct them to touch specified multiple corners at each step using two or three fingertips. A participant undergoes 23 steps of such touching processes per-hand. The Tab.~\ref{tab:handcalib} comprises hand calibration instructions for subjects to follow. Corner \# is a set of two or three target corner numbers of the calibration structure which the subject should contact with their fingertips. Also, the orders of fingers to touch the target corners are specified with numbers corresponding to each finger, which are (1:Thumb, 2:Index, 3:Middle, 4:Ring, 5:Little). An example of the hand calibration procedure is shown in our supplementary video.

\subsection{Implementation Details on Hand Calibration}

Here we describe the details of the hand calibration method. As described in section 3.4, optimization parameters are hand skeleton configuration $\mathcal{H}=\{\mathcal{S}^h, \mathcal{O}^h\}$ and positions of 3D markers in the local hand-centric coordinate $\mathcal{M}^h$. We empirically decide the general target range of the optimization skeleton to the palm area shown in Fig.~\ref{fig:supp_combined} and add constraints that limit the skeleton scales($s_i$) for each skeleton segment $i$ between $0.8\le s_i \le 1.2 $, and additional skeleton offset value($\delta_j$) for target joints $j$ with $\left| \delta \right| \le 0.01$ in meter scale to avoid unnatural deformation of hand skeleton. The location of hand markers $\mathcal{M}^h$ is optimized through a total of 150 iterations. The skeleton scale and additional offset are optimized starting from 50 and 100 iterations each. We use three losses, $L_{tip}$ to measure the Euclidean distance from the hand tips to paired corners, $L_{wrist}$ to measure the distance from the wrist location from body motion capture device and the wrist position computed by hand marker position and $L_{pen}$ to measure penetration of hand to the calibration structure. The penetration loss is computed by a cosine similarity between the calibration structure's normal vector and the target corner-to-hand tip vector. In summed loss $\lambda_{t} L_{tip} +  \lambda_{w}\mathcal{L}_{wrist} +\lambda_{p}\mathcal{L}_{pen}$, losses are weighted equally by $\lambda_{t}=1,\lambda_{w}=1, \lambda_{p}=1$. But they are manually adjusted based on the touch accuracy and body calibration accuracy per participant. After the alignment process, the average Euclidean distance between the corner and the target fingertip results in $0.83$ (in centimeters).

\begin{figure}[t!]
    \centering
    \includegraphics[width=\linewidth, trim={0cm 0.0cm 0cm 0.2cm},clip]{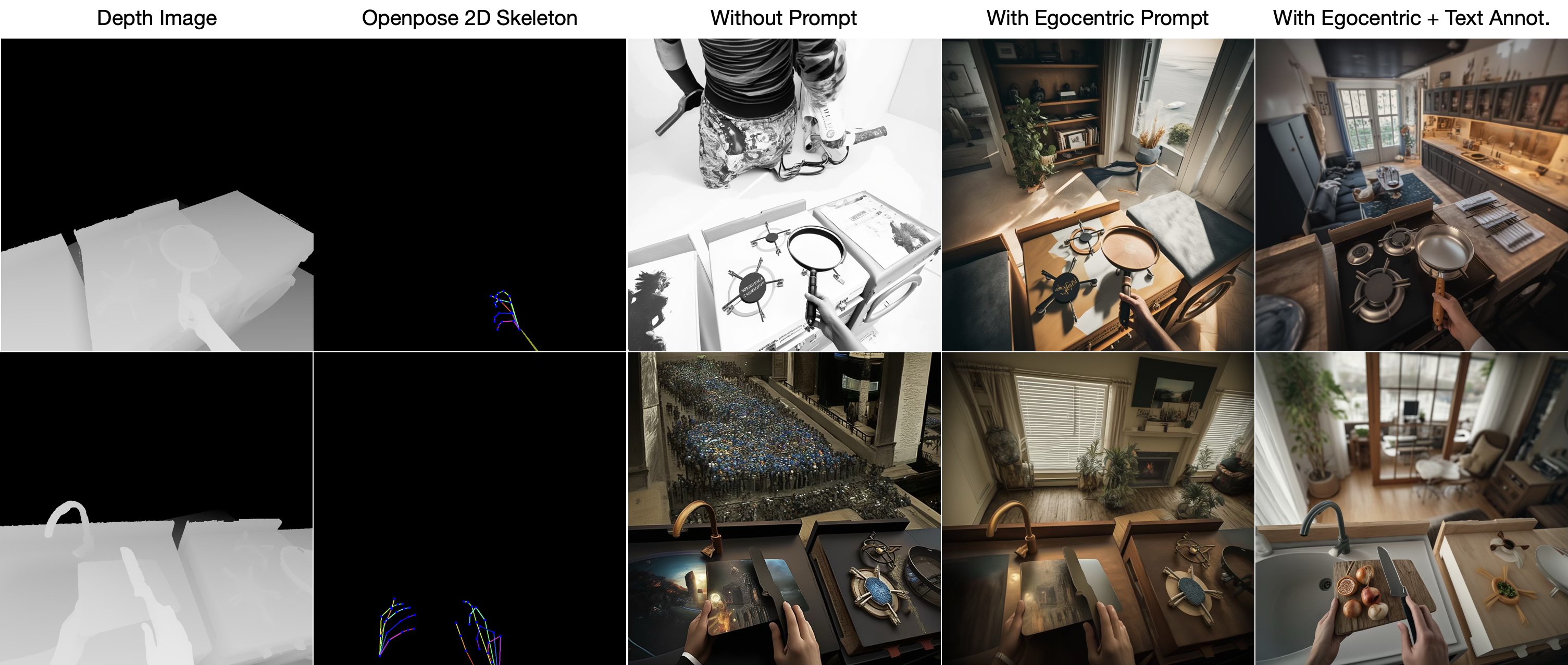}
    \caption{Synthesized RGBs and Comparison between with or without text annotation.}
    \label{fig:synthetic_annot_effect}
    \vspace{-10pt}
\end{figure}

\subsection{Fitting Human Body Model}
We illustrate details on fitting SMPL-X~\cite{mahmood2019amass} pose parameters to Xsens skeletons using the optimized shape parameters. For SMPL-X hand pose, we devise an optimization scheme which runs FABRIK solver~\cite{Aristidou:2016:ExtFABRIK} initially to get SMPL-X hand skeleton retargetted to Xsens hand and optimize each hand joint pose directly to fit into each retargetted joint positions. For body pose, we do not consider global orientation and translation in acquiring body pose for simplicity. As an input data representation, we split each sequence with a window size of 60, and reform body joint rotations except for hands in 6D representations~\cite{zhou20196d}, thus $x_{in} \in \mathbb{R}^{60 \times 21 \times 6}$ per batch. 
For the model, we use a variation of Temporal Convolution Network~\cite{lea2017temporal} for the encoder and decoder. During training, we define the default reconstruction loss, \(\mathcal{L}_{\text{recon}}\), for joint rotation and further incorporate the end effector loss, \(\mathcal{L}_{\text{end}}\). This additional loss includes the SMPL-X vertices of the hands and legs, as well as the wrist, foot, and hand tip joints. To regulate any present noise in the motion, we also add joint velocity loss as an regularization. Then the loss sums up to:

$$\mathcal{L} = \lambda_{recon} \mathcal{L}_{recon} + \lambda_{end} \mathcal{L}_{end} +\lambda_{vel} \mathcal{L}_{vel}$$

After training, we extract windows of all sequences with step size of 30, and initialize a latent code $z_{pose}$ by feeding the encoder with the Xsens joint rotation data by roughly matching joint category between two different skeletons(i.e. jLeftT4Shoulder of Xsens to left collar of SMPL-X). Then we optimize $z_{pose}$ by feeding into the trained decoder to fit with Xsens skeleton wrist, hand tip, ankle, and foot joints. Thus denoting a set of paired Xsens and SMPL-X target joints as $\mathcal{J}$, we formulate the optimization problem as:

$$
z_{\text{pose}}^* = \argmin_{z_{\text{pose}}} \sum_{j_{\text{xsens}},j_{\text{smplx}} \in \mathcal{J}} \| j_{\text{xsens}} - j_{\text{smplx}} \|^2
$$

After optimization, we use decoded output body pose using $z_{pose}^*$. Since we sample sequences as 60-length windows with step size of 30, there exists discrepancies in body poses where contiguous windows overlap. We use \textit{slerp} to compensate for such discontinuities for each joint pose parameters.

\subsection{Synthesizing Realistic RGB}

\begin{table}[tb]
  \centering
  \resizebox{\columnwidth}{!}{
    \begin{tabular}{c|c|ccc}
      \toprule
      \textbf{Dataset} & \textbf{Method} & \textbf{AUC@IoU$_{25}$}$\uparrow$ & \textbf{AUC@IoU$_{50}$}$\uparrow$ & \textbf{AUC@IoU$_{75}$}$\uparrow$ \\
      \midrule
       ROPE~\cite{zhang2025omni6d} & IST-Net~\cite{liu2023prior} & 28.7 & 10.6 & 0.5\\
       ROPE~\cite{zhang2025omni6d} & GenPose++~\cite{zhang2025omni6d} & \textbf{39.9} & \textbf{19.1} & \textbf{2.0}\\
       ParaHome$_{all}$ & GenPose++~\cite{zhang2025omni6d} & 26.4 & 10.3 & 0.6 \\
       ParaHome$_{rigid}$ & GenPose++~\cite{zhang2025omni6d} & 29.7 & 12.2 & 0.9 \\
    \bottomrule
    \end{tabular}
  }
  \caption{Quantitative comparison of category-level object pose estimation on ROPE~\cite{zhang2025omni6d} and \ParaHouse synthetic data. Since the two datasets differ in the presence of articulation, we divide \ParaHouse data into two subsets, $all$ including articulation objects and $rigid$ with only rigid objects.}
  \label{tab:synthetictable1}
  \vspace{-10pt}
\end{table}

\begin{table}[tb]
  \centering
  \resizebox{\columnwidth}{!}{
    \begin{tabular}{c|c|cccc}
      \toprule
      \textbf{Dataset} & \textbf{Method} & \textbf{PA-MPJPE$\downarrow$} & \textbf{PA-MPVPE$\downarrow$} & \textbf{F@5$\uparrow$} & \textbf{F@15$\uparrow$} \\
      \midrule
      FreiHAND~\cite{zimmermann2019freihand} & HaMeR~\cite{pavlakos2024reconstructing} & 6.0 & 5.7 & 0.785 & 0.990 \\
      HO3D~\cite{hampali2020honnotate} & Pose2Mesh~\cite{pavlakos2024reconstructing} & 12.5 & 12.7 & 0.441 & 0.909 \\
      HO3D~\cite{hampali2020honnotate} & HaMeR~\cite{pavlakos2024reconstructing} & 7.7 & 7.9 & 0.635 & 0.980 \\
      ParaHome(Ours) & HaMeR~\cite{pavlakos2024reconstructing} & 9.47 & 9.46 & 0.25 & 0.85 \\
    \bottomrule
    \end{tabular}
  }
  \caption{Quantitative comparison of 3D hand pose reconstruction on FreiHand, HO3D and \ParaHouse synthetic data. PA-MPVPE and PA-MPJPE are meausred in $mm$.}
  \label{tab:synthetictable2}
  \vspace{-10pt}
\end{table}

\begin{figure}[t!]
    \centering
    \includegraphics[width=\linewidth, trim={0cm 0.0cm 0cm 0.2cm},clip]{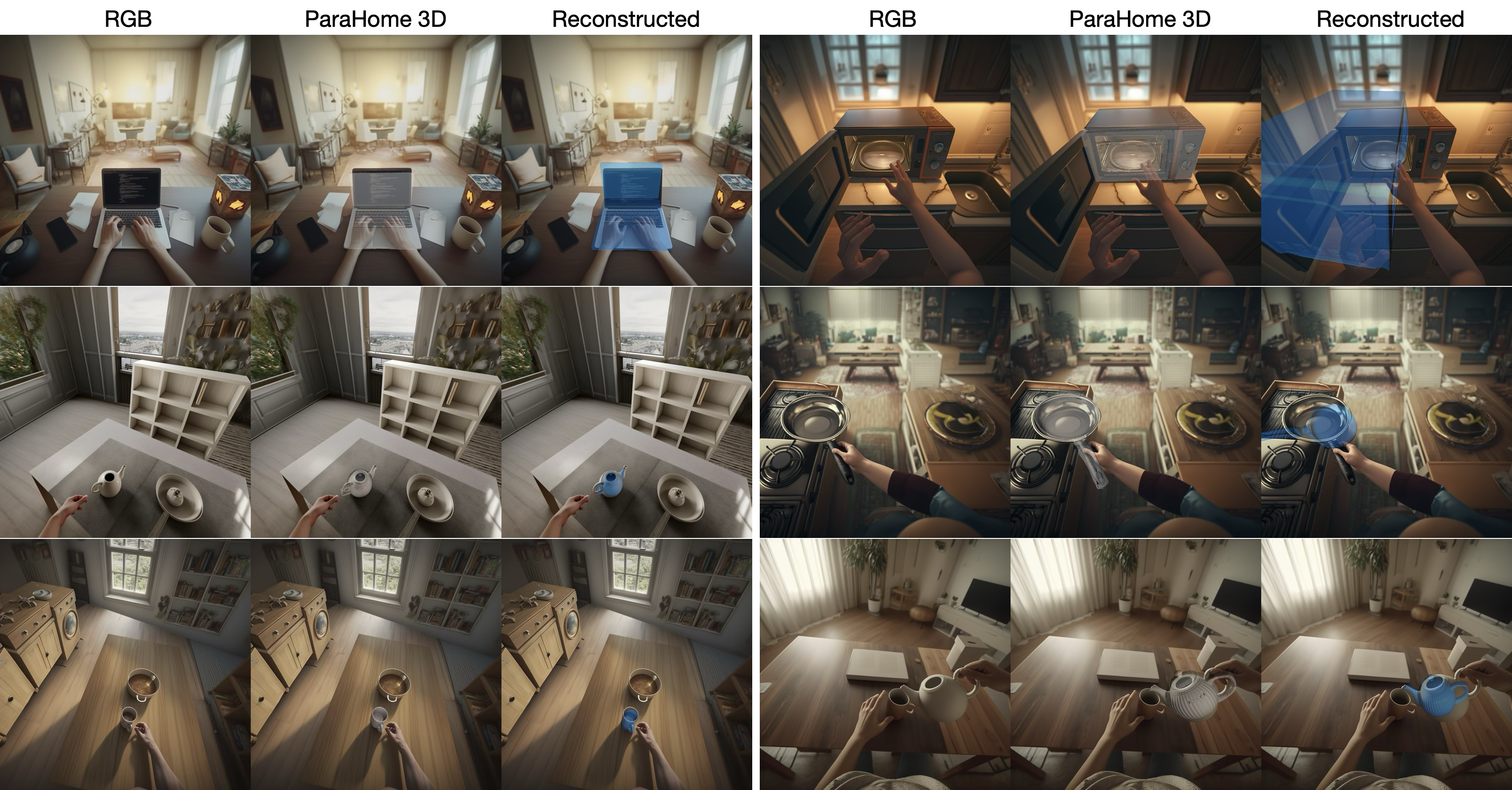}
    \caption{Rendered results of 6d reconstruction model on synthesized data. (Left) Successful cases. (2) Failure cases due to the occlusion.}
    \label{fig:supp_obj6d_quality}
    \vspace{-10pt}
\end{figure}

\begin{figure}[t!]
    \centering
    \includegraphics[width=\linewidth, trim={0cm 0.0cm 0cm 0.2cm},clip]{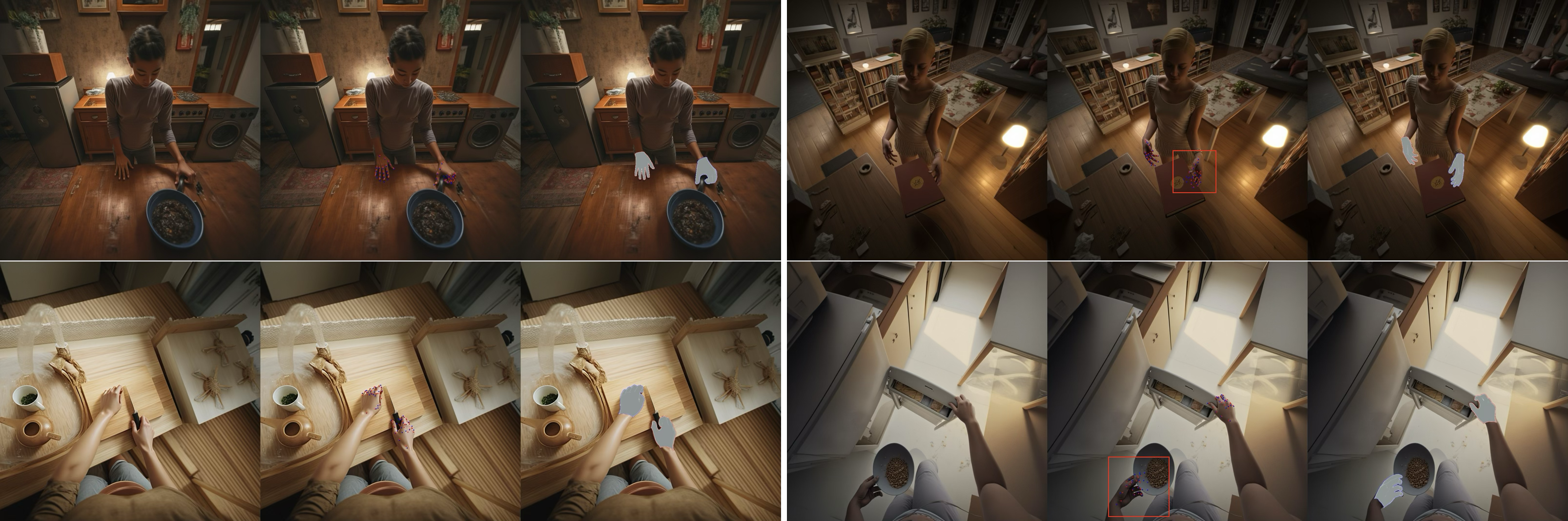}
    \caption{Rendered results of the 3D hand reconstruction model on synthesized data. (Left) Successful cases. (Right) Cases with large 3D keypoints loss due to occlusion.}
    \label{fig:supp_hand_quality}
    \vspace{-10pt}
\end{figure}

Utilizing the \ParaHouse dataset, which provides diverse and rich 3D motion data, we generate RGB images from various viewpoints, all aligned with 3D annotations. We employ a diffusion-based image synthesis model~\cite{flux} combined with ControlNet to create 2D RGB images consistent with the 3D data. Human-object interaction scenes are rendered from multiple perspectives, including egocentric, high-angle, and front-facing views. Rendered depth maps and OpenPose~\cite{cao2017realtime} joint information are integrated, along with text prompts to enhance image quality and alignment with the original data. The impact of including text information is shown in Fig.~\ref{fig:synthetic_annot_effect}, implying improved alignment and realism. Quantitative results on off-the-shelf 3D estimation models are presented in Table~\ref{tab:synthetictable1} and Table~\ref{tab:synthetictable2}. For object 6D estimation, our synthetic data achieves accuracy comparable to the ROPE dataset, demonstrating the realism of the generated images for model to detect. Specifically, occlusions from hand interactions and complex object articulation in the \ParaHouse dataset result in lower accuracy, as shown in Fig.\ref{fig:supp_obj6d_quality}, suggesting future potential improvement. For 3D hand pose estimation, the synthetic data performs competitively with other datasets quantitatively, though occlusions during manipulation lead to slightly reduced accuracy compared to HO3D\cite{hampali2020honnotate} and FreiHAND~\cite{zimmermann2019freihand}, as illustrated in Fig.~\ref{fig:supp_hand_quality}.

\section{Sequence Example Visualization} 
\subsection{Sequence Visualization}
\begin{figure*}[p]
    \centering
    \includegraphics[width=\linewidth, trim={0cm 0.0cm 0cm 0cm},clip]{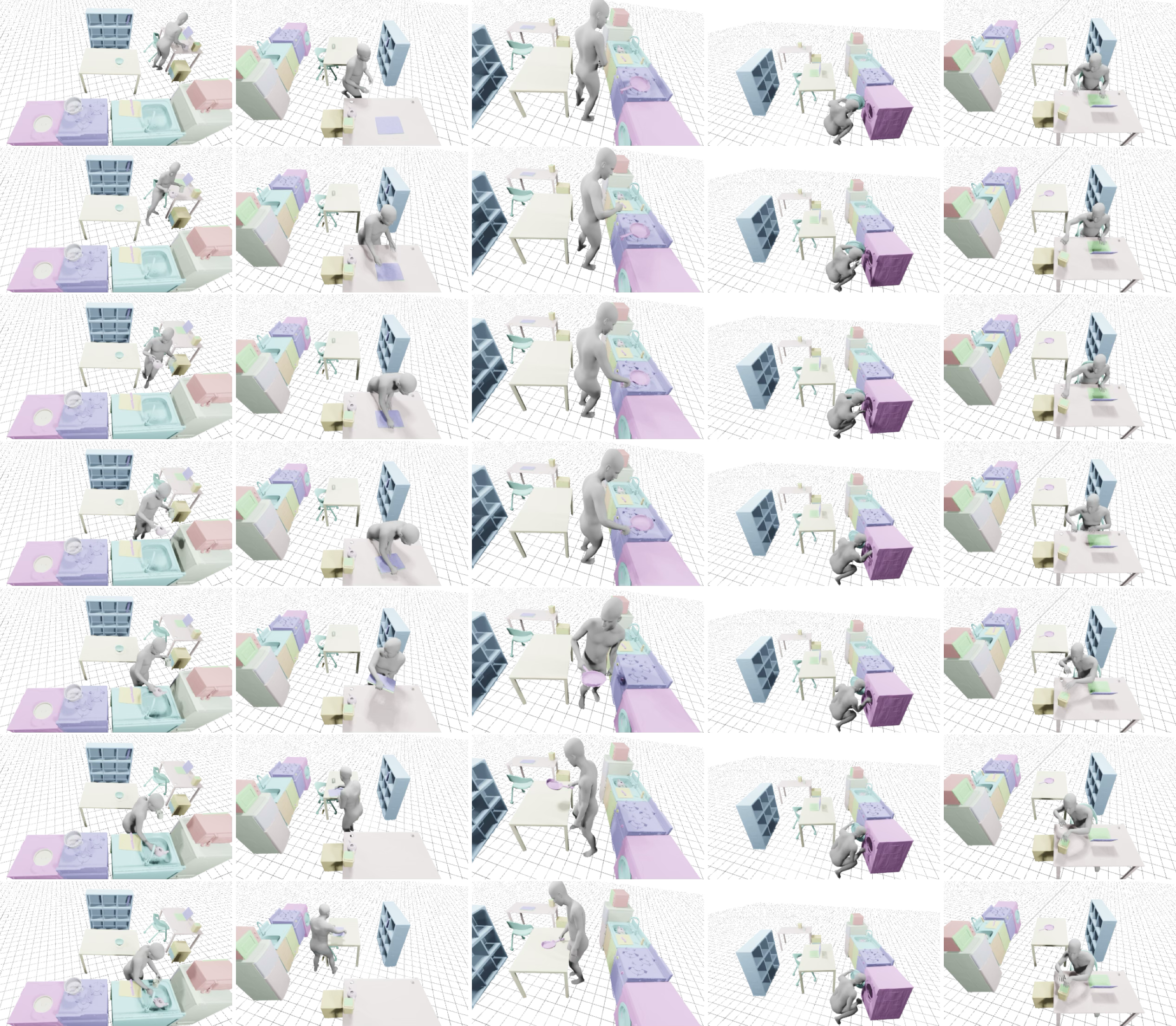}
    \caption{Example scenes of ParaHome dataset and aligned text annotation (Column1) Move kettle and cup from desk to the sink. (Column2) Take laptop from the desk and move to the table. (Column3) Take pan from the gas stove to the table. (Column4) Put laundry in the washing machine. (Column 5) Throw away STH into trash can.}
    \label{fig:scene_example}
    \vspace{-10pt}
\end{figure*}

Sampled data from our collected datasets are shown in Fig.~\ref{fig:scene_example}. Corresponding text annotations for actions are provided under the caption.

\section{Experiments}

\subsection{Synthesizing Body Motion for Desired Object Manipulation}

\begin{figure}[t!]
    \centering
    \includegraphics[width=\linewidth, trim={0cm 0.0cm 0cm 0.2cm},clip]{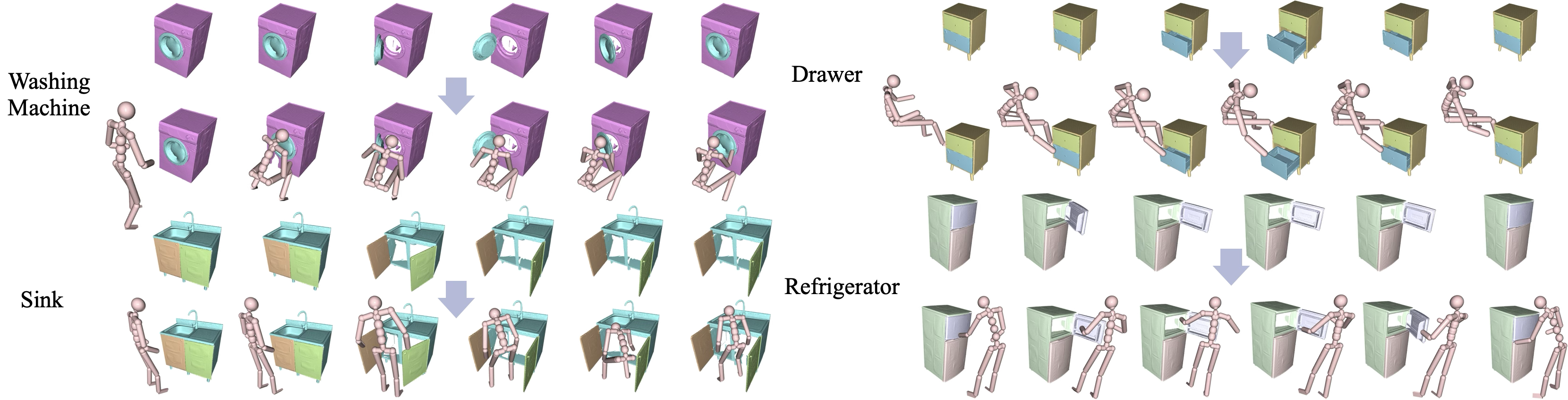}
    \caption{Synthesized body motions conditioned by sequences of object state}
    \label{fig:supp_task1_result}
    \vspace{-10pt}
\end{figure}

\begin{figure}[tb]
    \centering
    \includegraphics[width=\linewidth, trim={0cm 0.0cm 0cm 0cm},clip]{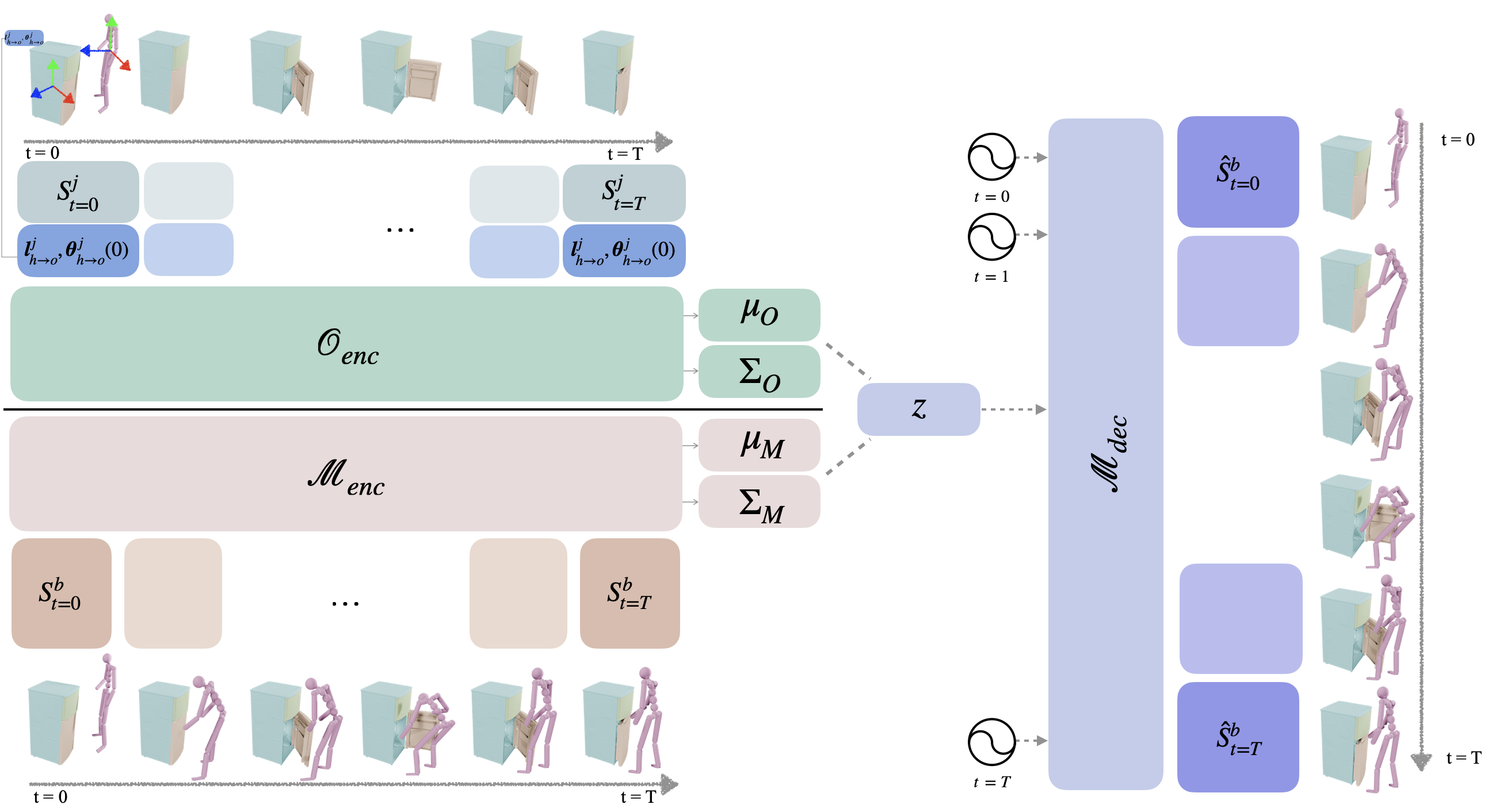}
    \caption{Model Architecture estimating human motion based on object states}
    \label{fig:supp_task1}
    \vspace{-10pt}
\end{figure}

\noindent \textbf{Train Details}: The goal of our model shown in Fig.~\ref{fig:supp_task1} is to synthesize a plausible 3D human motion conditioned with sequences of object state at a range of times. We represent target object status at each time $t$ as $\mathbf{S}_{to}(t)=\{\bs{\phi}^j(t)\}$ using joints state $\bs{\phi}^j(t)\in \mathbb{R}^2$. We represent body pose at time $t$ as $\mathbf{S}_p(t) = \{ X_t, \Delta{p}, \Delta{r}\}$ using body pose $X_t$, root's linear velocity $\Delta{p}$ and angular velocity$\Delta{r}$. We test with two types of body pose representation: the person root-centered skeleton representation~\cite{holden2016deep}, and the SMPL-X~\cite{pavlakos2019expressive} body pose. In training, we use AdamW optimizer and LR=$1e-4$ with $1500$ epochs and batch size 32.

\begin{table}[tb]
  \scriptsize
  \centering
  \resizebox{\columnwidth}{!}{
  \begin{tabular}{c|c|c|cccc|c|c}
    \toprule
    \multicolumn{1}{c}{} & \multicolumn{1}{c}{} & \multicolumn{1}{c}{} &  \multicolumn{4}{c}{\textbf{MPE$\downarrow$}(cm)} & \multicolumn{1}{c}{} & \multicolumn{1}{c}{}\\
    \cmidrule(rl){4-7} 
    Object & body repr. & window & rc-joints & rc-wrists & glb-root & glb-joints & \textbf{MOE$\downarrow$} &
    \textbf{Foot skating}$\downarrow$  \\
    \midrule
     \multirow{6}{*}{Refrigerator} 
     &  \multirow{3}{*}{root centric} 
     & 30 & 0.58 & 1.36 & 0.68 & 1.15 & 0.22 & 0.98\\
     && 60 & 0.76 & 1.73 & 2.47 & 3.22 & 0.43 & 1.05\\
     && 90 & 0.97 & 2.09 & 5.60 & 6.57 & 0.75 & 1.74\\
     &  \multirow{3}{*}{SMPL-X} 
     & 30 & 1.67 & 2.69 & 0.68 & 2.32 & 0.12 & 1.58\\
     && 60 & 1.93 & 2.94 & 1.83 & 3.95 & 0.24 & 1.50\\
     && 90 & 3.49 & 5.26 & 5.32 & 8.87 & 0.53 & 2.67\\
    \midrule
     \multirow{6}{*}{Drawer} 
     &  \multirow{3}{*}{root centric} 
     & 30 & 1.51 & 2.10 & 0.81 & 1.86 & 0.13 & 1.04\\
     && 60 & 1.52 & 1.74 & 1.02 & 2.26 & 0.24 & 0.80\\
     && 90 & 2.61 & 2.90 & 3.92 & 5.88 & 0.66 & 1.58\\
     &  \multirow{3}{*}{SMPL-X} 
     & 30 & 3.43 & 4.04 & 0.40 & 3.87 & 0.11 & 1.38\\
     && 60 & 3.27 & 3.83 & 0.96 & 4.17 & 0.12 & 1.04\\
     && 90 & 2.78 & 3.39 & 2.73 & 5.20 & 0.18 & 1.10\\
     \midrule
     \multirow{6}{*}{Washing Machine} 
     &  \multirow{3}{*}{root centric} 
     & 30 & 0.50 & 0.89 & 0.35 & 0.77 & 0.15 & 0.94\\
     && 60 & 0.55 & 1.06 & 1.44 & 2.14 & 0.48 & 1.35\\
     && 90 & 1.29 & 2.62 & 9.03 & 11.46 & 0.92 & 2.60\\
     &  \multirow{3}{*}{SMPL-X} 
     & 30 & 1.94 & 2.53 & 0.42 & 2.24 & 0.09 & 1.04\\
     && 60 & 3.68 & 4.29 & 2.53 & 6.59 & 0.44 & 2.72\\
     && 90 & 6.03 & 7.55 & 6.01 & 13.29 & 0.60 & 2.68\\
      \midrule
     \multirow{6}{*}{Sink} 
     &  \multirow{3}{*}{root centric} 
     & 30 & 0.58 & 0.98 & 0.42 & 0.82 & 0.16 & 0.79\\
     && 60 & 0.61 & 0.94 & 1.02 & 1.47 & 0.29 & 0.81\\
     && 90 & 1.21 & 1.93 & 3.83 & 4.84 & 0.76 & 1.67\\
     &  \multirow{3}{*}{SMPL-X} 
     & 30 & 2.27 & 2.91 & 0.52 & 2.64 & 0.10 & 0.90\\
     && 60 & 2.64 & 3.12 & 1.26 & 3.79 & 0.18 & 0.87\\
     && 90 & 2.60 & 3.12 & 2.16 & 4.80 & 0.27 & 1.01\\
  \bottomrule
  \end{tabular}
  }
  \caption{Quantitative results of the \ParaHouse task.}
  \label{tab:quantitative_benchmark1_supp}
\end{table}

\noindent \textbf{Additional Results}: We train our baseline model for four objects including a refrigerator, drawer, washing machine, and sink with window sizes 30, 60, and 90. The quantification results are shown in Table~\ref{tab:quantitative_benchmark1_supp}. As shown in the result table, as the window size decreased, the accuracy increased in most items and root-centered body skeleton representation results in better accuracy for pose-dependent attributes (rc-joints, rc-wrists, glb-joints) but SMPL-X notation results in better global orientation and root position. Additional examples of visualization are shown in Fig.~\ref{fig:supp_task1_result} and our supplementary video.

\end{document}